%% file: _main.tex
\documentclass{article} 
\usepackage{wrapfig}
\usepackage{booktabs}
\usepackage[round]{natbib}

\input{math_commands.tex}

\input{arxiv_setting}

\usepackage{placeins} 
\usepackage{afterpage} 

\usepackage{url,hyperref}
\usepackage{graphicx, float, subcaption, xcolor}
\usepackage{kotex}

\definecolor{darkblue}{rgb}{0.0,0.0,0.65}
\definecolor{darkred}{rgb}{0.65,0.0,0.0}
\definecolor{darkgreen}{rgb}{0.0,0.5,0.0}
\definecolor{tab:blue}{RGB}{31,119,180}  
\definecolor{tab:red}{RGB}{214,39,40}  
\definecolor{tab:green}{RGB}{44,160,44}  
\definecolor{tab:orange}{RGB}{255,127,14}  
\hypersetup{
	colorlinks = true,
	citecolor  = darkblue,
	linkcolor  = darkred,
	filecolor  = darkblue,
	urlcolor   = darkblue,
}

\newcommand{\kunwoo}[1]{\textcolor{darkred}{[Kunwoo: #1]}}

\usepackage[normalem]{ulem}

\usepackage[margin=1.2in]{geometry}
\usepackage{latexsym, amsthm, amscd, amsfonts, mathrsfs, amsmath, amssymb, stmaryrd, tikz-cd, mathrsfs, bbm, url, esint, mathtools, enumerate}
\usepackage{blkarray}
\usepackage{dsfont}

\usepackage{authblk}

\title{Probability-Flow ODE in Infinite-Dimensional Function Spaces}

\author[1]{Kunwoo Na\thanks{This work was done while Kunwoo Na was an intern at KAIST AI}}
\author[2]{Junghyun Lee}
\author[2]{Se-Young Yun}
\author[3,4]{Sungbin Lim}

\affil[1]{Seoul National University~(SNU)}
\affil[2]{Korea Advanced Institute of Science and Technology~(KAIST)}
\affil[3]{Korea University}
\affil[4]{LG AI Research}

\affil[ ]{\small \texttt{jamongna@snu.ac.kr}, \texttt{\{jh\_lee00,yunseyoung\}@kaist.ac.kr}, \texttt{sungbin@korea.ac.kr}}

\date{}

\usepackage[parfill]{parskip}

\begin{document}

\maketitle
\begin{abstract}
    Recent advances in infinite-dimensional diffusion models have demonstrated their effectiveness and scalability in function generation tasks where the underlying structure is inherently infinite-dimensional. 
    To accelerate inference in such models, we derive, for the first time, an analog of the probability-flow ODE~(PF-ODE) in infinite-dimensional function spaces. Leveraging this newly formulated PF-ODE, we reduce the number of function evaluations while maintaining sample quality in function generation tasks, including applications to PDEs.
\end{abstract}


\input{section/01}
\input{section/02}

\input{section/03}
\input{section/04}

\subsubsection*{Acknowledgments}
The authors thank Hojung Jung for providing helpful feedback on the initial manuscript.
This work was supported by the Institute of Information \& Communications Technology Planning \& Evaluation (IITP) grant funded by the Korean government~(MSIT) (No.RS-2019-II190075 Artificial Intelligence Graduate School Program (KAIST), No.2022-0-00612, Geometric and Physical Commonsense Reasoning based Behavior Intelligence for Embodied AI, No.RS-2022-II220311, Development of Goal-Oriented Reinforcement Learning Techniques for Contact-Rich Robotic Manipulation of Everyday Objects), and the National Research Foundation of Korea~(NRF) grant funded by the Korea government~(MSIT) (RS-2024-00410082).

\newpage
\bibliography{references}
\bibliographystyle{plainnat}
\nocite{*}

\appendix
\input{section/A01}

\input{section/A02}
\input{section/A03}
\end{document}

%% file: math_commands.tex
\usepackage{amsmath,amsfonts,bm,amssymb}
\usepackage{dsfont}
\usepackage{mathrsfs}  

\def\eqref#1{Eqn.~(\ref{#1})}

\def\1{\bm{1}}

\DeclareMathAlphabet{\mathsfit}{\encodingdefault}{\sfdefault}{m}{sl}
\SetMathAlphabet{\mathsfit}{bold}{\encodingdefault}{\sfdefault}{bx}{n}

\def\gH{{\mathcal{H}}}

\def\gL{{\mathcal{L}}}
\def\gM{{\mathcal{M}}}

\DeclareMathOperator*{\minimize}{minimize}
    
    \DeclareMathOperator*{\expectation}{\mathbb{E}}

    \newcommand{\NN}{\mathbf N}
    \newcommand{\PP}{\mathbf P}
    
    \newcommand{\RR}{\mathbf R}

\usepackage{amsthm}
\newtheorem{thm}{Theorem}[section]
\newtheorem*{theorem*}{Theorem}
\newtheorem{theorem}[thm]{Theorem}

\newtheorem{lemma}[thm]{Lemma}

\newtheorem*{claim*}{Claim}

\newtheorem*{assumption*}{Assumption}

\newcommand{\thmref}[1]{Theorem~\ref{#1}}

\newcommand{\lemref}[1]{Lemma~\ref{#1}}

\newtheorem{definition}[thm]{Definition}
\newtheorem*{definition*}{Definition}
\newtheorem*{remarks*}{Remarks}
\newtheorem*{remark*}{Remark}

%% file: arxiv_setting.tex
\usepackage[OT4]{fontenc}
\usepackage[margin=1.2in]{geometry}
\usepackage{amssymb,amsfonts,amsmath,amsthm,amscd,dsfont,mathrsfs,pifont}
\usepackage{blkarray}
\usepackage{graphicx,float,psfrag,epsfig,color}
\usepackage{qtree}
\usepackage{comment}
\usepackage{authblk,textcomp}
\usepackage{caption} 
\usepackage{subcaption}

 \usepackage{dsfont}
\usepackage{blkarray}
\usepackage[utf8]{inputenc} 
\usepackage[T1]{fontenc}    
\usepackage{hyperref}       
\usepackage{url}            
\usepackage{booktabs}       
\usepackage{nicefrac}       
\usepackage{microtype}      

\usepackage{algpseudocode}
\usepackage{enumitem}
\usepackage{comment}
\usepackage{bbm}
\usepackage[ruled,vlined]{algorithm2e}

%% file: section/01.tex
\section{Introduction}

Diffusion model~\citep{sohl2015deep, ho2020denoising, song2021score, kingma2021variational} is a class of generative model that adds noise to real data to train the score network and sequentially approximate the time-reversed process~\citep{follmer1986time, anderson1982reverse} to generate samples from the true data distribution.
This model has shown remarkable empirical success in numerous domains such as image generation~\citep{song2021score, song2021maximum}, video generation~\citep{luo2023videofusion}, medical data processing~\citep{song2022solving, chung2022score, akrout2023diffusion}, and audio generation~\citep{kongdiffwave}.

However, ``classical" diffusion models formulated on finite-dimensional Euclidean spaces limit their applicability to function generation problems as they can only generate function values realized on a fixed discretization of the function's domain~\citep{li2020fourier} and cannot capture functional properties of a data such as integrability or smoothness~ \citep{kerrigan2023diffusion}.
Motivated by such a limitation of finite-dimensional models, there has been a line of works extending the finite-dimensional diffusion model to infinite-dimensional Hilbert spaces; for instance, \cite{hagemann2023multilevel, kerrigan2023diffusion, jlim2023score-based, lim2023score, pidstrigach2023infinite, phillips2022spectral, baldassari2023conditional}. 
\cite{kerrigan2023diffusion} proposes a discrete-time model that serves as an analog of \cite{ho2020denoising} in infinite-dimensional space, and \cite{hagemann2023multilevel} introduces a finite-dimensional approximation of an infinite-dimensional SDEs and utilizes the time-reversal formula in finite-dimensional spaces. \cite{jlim2023score-based, franzese2023continuoustime, pidstrigach2023infinite} propose continuous-time models by extending the SDE framework of \cite{song2021score} to infinite dimensions based on semigroup theory~(ref. \citet{da2014stochastic}); however, their consideration is limited to a relatively simple class of SDEs, such as Langevin type SDE or SDEs with constant-time diffusion coefficients. Later, \cite{lim2023score} proved a general form of time-reversal formula which encompasses various choices of SDEs such as VPSDE, VESDE, sub-VPSDE~\citep{song2021score} and variance scheduling~\citep{nichol2021improved}, by exploiting more advanced mathematical machinery, e.g., variational approach and functional derivatives~(ref. \citet{krylov2007stochastic, bogachev1999absolutely}).

Research works mentioned above are primarily focused on the \textit{training} of diffusion models, i.e., they aim to implement a mathematical framework in which the score-matching objective~\citep{sohl2015deep, vincent2011connection} and time reversal~\citep{follmer1986time, millet1989time} of the noising process are possible. Although the ``SDE'' component and the ``score-matching'' component of the finite-dimensional diffusion model have been transferred to infinite dimensions, the existence of an infinite dimensional analog of {\bf probability-flow ODE}~(PF-ODE; \cite{song2021score}) is still open. Indeed, PF-ODE has been crucial in the sampling process of diffusion models as it allows for fast sampling~\citep{chen2023pfode, lu2022dpm} and consistency modeling~\citep{song2023consistency}. In this work, we aim to accelerate the \textit{inference} process of infinite-dimensional diffusion models by extending the probability-flow ODE~\citep{song2021score} to infinite-dimensional spaces.

\paragraph{Contributions.}
Our contributions are as follows: 
\begin{itemize}
    \item We derive in a mathematically rigorous manner the notion of probability-flow ODE~(Theorem~\ref{thm:prob-flow-ode}) associated with a general class of stochastic differential equations~(SDEs) 
    in infinite-dimensional spaces, including VPSDE, VESDE, sub-VPSDE~\citep{song2021score} and variance scheduling~\citep{nichol2021improved}. We note that our infinite-dimensional probability-flow ODE is widely applicable regardless of the specific formulation of the infinite-dimensional diffusion model. 

    \item We empirically demonstrate that sampling with PF-ODE achieves comparable or superior generation quality to the previous SDE-based approach while requiring significantly fewer number of function evaluations~(NFEs) in both toy and real-world PDE problems.
\end{itemize}

\section{Preliminaries}

\subsection{Probability-flow ODE in $\RR^n$}

Let us consider the following stochastic differential equation in $\RR^n$ ($n < \infty$) over $t \in [0, T]$:
\begin{equation}
    dX_t = f( t, X_t) dt + \sigma(t) dB_t, \quad X_0 \sim p_0 = p_\mathrm{data}, 
\end{equation}
where $(B_t)_{t \ge 0}$ is a standard Brownian motion in $\RR^n$, $f : [0, T]  \times \RR^n \rightarrow \RR^n$ is the drift term, $\sigma : [0, T] \rightarrow \operatorname{Mat}_n(\RR)$ is the diffusion term, and $p_0 = p_\mathrm{data}$ is the probability \textit{density} of the target data distribution. Closely related to this SDE is the so-called \textbf{probability-flow ODE} (PF-ODE; \cite{song2021score}):
\begin{align*}
    dY_t = \left[
    f(t,Y_t)  -\frac{1}{2} A(t) \nabla  \log p_t(Y_t) 
    \right]dt, \quad Y_0 \sim p_0,
\end{align*}
where $A(t) = \sigma(t) \sigma(t)^\intercal$, and $p_t$ is the density of $X_t$. 
It is well-known that the solution for the PF-ODE has the same density as $X_t$ for each $t$~\citep[Appendix D.1]{song2021score}. The derivation of the PF-ODE heavily relies on the Fokker-Planck equation~(ref. \cite{oksendal2003stochastic} for example), a well-studied second-order PDE whose solution is $(t, x) \mapsto p_t(x)$. In infinite-dimensional spaces, however, one cannot utilize the probability density function in the analysis due to the lack of reference measure~(ref. \citet{lunardi2015infinite}, Proposition 2.2.1). Hence, a more careful treatment is required for infinite-dimensional cases.

\subsection{Infinite-Dimensional Analysis}

Let $\mathcal{H}$ denote a real separable Hilbert space, and $(W_t)_{t \ge 0}$ be a $Q$-Wiener process on $\mathcal{H}$. Denote by $\mathcal{H}_Q$ the Cameron-Martin space~(ref. \cite{da2014stochastic}) of $\mathcal{N}(0, Q)$. 
Let $\mathscr{L}_2(\mathcal{H})$ be the set of Hilbert-Schmidt operators on $\mathcal{H}$, and let $\{\varphi_i\}$ be an orthonormal basis of $\mathcal{H}$ that consists of eigenvectors of $Q$ corresponding to $\lambda_i$. We assume $\mathcal{H}$ is a function space over some set $\Omega \subseteq \RR^d$~($d <\infty$); for example, $\mathcal{H } = L^2(\Omega)$ or $\mathcal{H} = W^{1, 2}(\Omega)$.

Due to the lack of reference measure in $\mathcal{H}$, we shall express the time evolution of a family of probability measures in a \textit{weak sense}; that is, we express the evolution of the dual pairings of a probability measure and test functions. Below, we introduce the minimal background required for this work; we refer the readers to Appendix~\ref{app:prelim} for a more detailed overview.

\paragraph{Test functions.} The class of cylindrical functions 
$\mathcal{FC}_b^\infty(\mathcal{H})$ is defined as 
\begin{align*}
    \mathcal{FC}_b^\infty(\mathcal{H}) = \left\{ x \mapsto 
   f(\left<\varphi_1, x\right>, \cdots, \left<\varphi_m, x \right>) \mathrel{\bigg|} m \in \NN, \ f \in \mathcal{C}_0^\infty(\RR^m)
    \right\}. 
\end{align*}
We write $  f_{\varphi_1, \cdots, \varphi_m}(x) =  f(\left<\varphi_1, x\right>, \cdots, \left<\varphi_m, x \right>)$ for $x \in \mathcal{H}$. Here, $\mathcal{C}_0^\infty(\RR^m)$ is the space of smooth functions on $\RR^m$ that vanish at infinity, which serves as a canonical class of test functions in usual finite-dimensional analysis.

\paragraph{Weak formulation.} Let $\mathcal{L}$ be an operator such that $\gL \psi: \mathcal{H} \to \RR$ is in $L^1(\mathcal{H}, \mu)$ for all $\psi \in \mathcal{FC}_b^\infty(\mathcal{H})$. We say $\gL^\ast \mu = 0$ if
\begin{align*}
    \int_\mathcal{H} \gL f_{\varphi_1, \cdots, \varphi_m}(x) \mu(dx) = 0, \quad \forall  f_{\varphi_1, \cdots, \varphi_m}\in \mathcal{FC}_b^\infty(\mathcal{H}). 
\end{align*}
In a similar manner, for a family of measures $\{\nu_t\}$, we shall understand the equation $ \mathcal{L}^\ast \mu = \partial_t \nu_t $
in a \textit{weak sense}, i.e., we say $\mathcal{L}^\ast \mu = \partial_t\, \nu_t$ if 
\begin{align*}
     \int_\mathcal{H} \gL f_{\varphi_1, \cdots, \varphi_m}(x) \mu(dx) = \frac{\partial}{\partial t} \int_\mathcal{H}  
       f_{\varphi_1, \cdots, \varphi_m}(x) \nu_t(dx),
     \quad \forall  f_{\varphi_1, \cdots, \varphi_m}\in \mathcal{FC}_b^\infty(\mathcal{H}).
\end{align*}

\paragraph{Logarithmic gradient.} 
We say that a Borel probability measure $\mu$ is Fomin differentiable along $h \in \mathcal{H}_Q$ if there exists a function $\rho_h^\mu \in L^1(\mathcal{H}, \mu)$ such that 
    \begin{align}
        \int_\mathcal{H} \partial_h f_{\varphi_1, \cdots, \varphi_m}(x) \mu(dx) = -\int_\mathcal{H} f_{\varphi_1, \cdots, \varphi_m}(x)  \rho_h^\mu (x) \mu(dx), \quad \forall  f_{\varphi_1, \cdots, \varphi_m}\in \mathcal{FC}_b^\infty(\mathcal{H}).
    \end{align}
    Here, $\partial_h f_{\varphi_1, \cdots, \varphi_m} (x)$ denotes the G\^ateaux differential of $f_{\varphi_1, \cdots, \varphi_m}$ at $x$ along $h$. 
    If there exists a function $\rho_{\mathcal{K}}^\mu : \mathcal{H} \to \mathcal{H}$ such that $\langle \rho_{\mathcal{K}}^\mu(x), h \rangle_{\mathcal{K}} = \rho_h^{\mu}(x)$
    for every $x \in \mathcal{H}$ and $h \in \mathcal{K}$, then we call $\rho_\mathcal{K}^\mu$ the logarithmic gradient of $\mu$ along $\mathcal{K}$. 

%% file: section/02.tex

\section{Probability-Flow ODEs in Function Spaces }

Let us consider an SDE in $\mathcal{H}$ given by 
\begin{align}
    dX_t = B(t, X_t) dt + G(t) dW_t, \quad X_0 \sim \PP_0 = \PP_\mathrm{data}, 
    \label{eqn:SDE-H}
\end{align}
where $(W_t)_{t \ge 0}$ is a $Q$-Wiener process on $\mathcal{H}$, $B: [0, T] \times \mathcal{H}\to \mathcal{H}$ and $G: [0, T] \to \mathscr{L}_2(\mathcal{H})$ are progressively measurable, and $\PP_0 = \PP_\mathrm{data}$ is the probability {\it measure} from which $X_0$ is sampled. Prior works~\citep{hagemann2023multilevel, lim2023score, jlim2023score-based, pidstrigach2023infinite} on infinite-dimensional diffusion models directly implement \eqref{eqn:SDE-H} and its time-reversal. On the other hand, in finite-dimensional models, PF-ODE has played a crucial role in allowing for faster sampling~\citep{lu2022dpm} and recently, leading to consistency modeling~\citep{song2023consistency}. 
Thus, it is only natural to ask whether there is an infinite-dimensional version of the PF-ODE, which, to the best of our knowledge, has not been tackled in the literature yet.

The usual approach of \citet{song2021score} of deriving the PF-ODE \textit{fails} in infinite-dimensions, as there is no probability density function. Our main question in this section is as follows:

\begin{center}
    \textit{
    Is there an ODE in infinite-dimensional space with a random initial point $Y_0 \sim \PP_0$ whose solution evolves like the solution of the original SDE~(\eqref{eqn:SDE-H})?
    }
\end{center}
The answer is affirmative.

Consider the following family of operators $\{\mathcal{L}_t\}_{t \in (0,T]}$ defined by 
\begin{align*}
        \mathcal{L}_t f_{\varphi_1, \cdots, \varphi_m}(u) = \frac{1}{2} \operatorname{Tr}_{\mathcal{H}_Q} \left(
        A(t) \circ Q \circ D^2 f_{\varphi_1, \cdots, \varphi_m}(u)
        \right) + \left<
        D f_{\varphi_1, \cdots, \varphi_m}(u), B(t, u) \right>_{\mathcal{H}_Q}
    \end{align*}
for $f_{\varphi_1, \cdots, \varphi_m} \in \mathcal{FC}_b^\infty(\mathcal{H}) $, where $A(t) = G(t) G(t)^\ast$ and $D$ stands for the Fr\'echet derivative.
It is known~(ref. \cite{belopolskaya2012stochastic}, Chapter 5) that for the solution of \eqref{eqn:SDE-H} denoted $X_t$, the law $\mu_t = \operatorname{Law}(X_t)$ satisfies the following \textit{Fokker-Planck-Kolmogorov} equation 
\begin{align*}
        \left\{ 
        \begin{aligned}
             &\partial_t {\mu}_t = (\mathcal{L}_t)^\ast {\mu}_t, \quad t \in (0, T], \\ 
             & {\mu}_t\Big|_{t = 0 } = \PP_0.
        \end{aligned}
      \right.
\end{align*}
Exploiting the preceding Fokker-Planck-Kolmogorov equation, we  explicitly state the PF-ODE in infinite-dimensional spaces as in the following theorem,
whose proof is deferred to Appendix~\ref{app:proof-pfode}:
\begin{theorem}
\label{thm:prob-flow-ode}
    Let $X_t$ be a solution of \eqref{eqn:SDE-H} and $\mu_t := \mathrm{Law}(X_t)$.
    Then, $\mu_t$ satisfies the Fokker-Planck-Kolmogorov equation of $(Y_t)_{t \in [0, T]}$, where $(Y_t)_{t \in [0, T]}$ is a solution of the following {probability-flow ODE} in infinite-dimension:
    \begin{align}
       dY_t = \left[ 
       B(t, Y_t)-  \frac{1}{2} A(t) \rho_{\mathcal{H}_Q}^{\mu_t}(Y_t)
       \right] dt, \quad Y_0 \sim \PP_0.
       \label{eq:1}
    \end{align}
    Here, $A(t) := G(t) G(t)^\ast$ and $\rho_{\mathcal{H}_Q}^{\mu_t}$ is the logarithmic gradient of $\mu_t$ along $\mathcal{H}_Q$.
\end{theorem}

%% file: section/03.tex
\vspace{4mm}
\section{Experiments} 

In all experiments, we sample synthetic functions via our PF-ODE and the usual time-reversed SDE in infinite-dimensional function spaces, where we employ the Euler's method for the ODE and SDE solving for each NFE. In Appendix~\ref{app:Experiments}, we provide the missing implementation details.

\subsection{1D Function generation}

\begin{wrapfigure}{r}{0.41\textwidth}
  \begin{center}    
  \vspace{-4mm}
\includegraphics[width=0.8\linewidth]{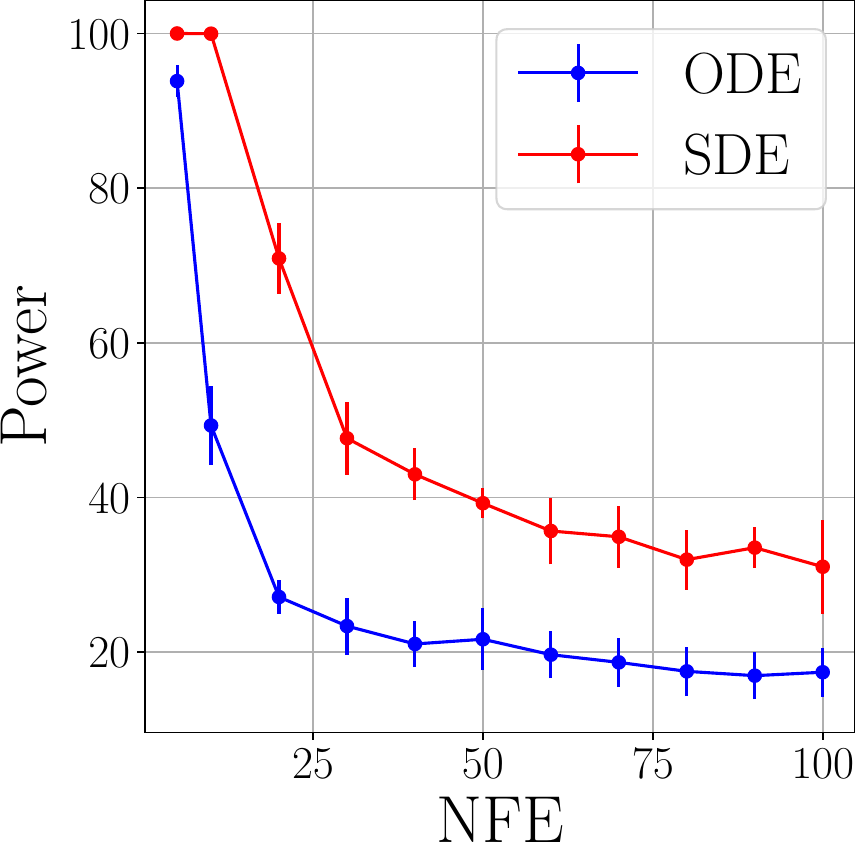}
  \end{center}
    \caption{Power vs. NFE.}
    \vspace{-17mm}
    \label{fig:Quadratic-NFE}
\end{wrapfigure}

\paragraph{Setting.}
We use a synthetic dataset $\texttt{Quadratic}$ consisting of (noise-corrupted) functions of the form $$f(x; a) = ax^2 + \varepsilon,$$ where $a \sim \mathrm{Unif}\{-1, 1\}$ and $\varepsilon \sim \mathcal{N}(0, 1)$ are sampled independently.
These functions are evaluated at a fixed grid $x = \texttt{np.linspace(-10, 10, 100)}$.
We utilize the checkpoint trained on the \texttt{Quadratic} dataset by \citet{lim2023score}.
For the evaluation, we calculate the power of kernel two-sample test with functional PCA kernel~(\cite{wynne2022kernel}; lower power is better). 
We consider the number of function evaluations~(NFEs) in the range of $\{10, 20, \cdots, 100\}$.

\begin{figure}[!ht]
\centering
\begin{subfigure}{0.31\textwidth}
\centering
\includegraphics[width=\textwidth]{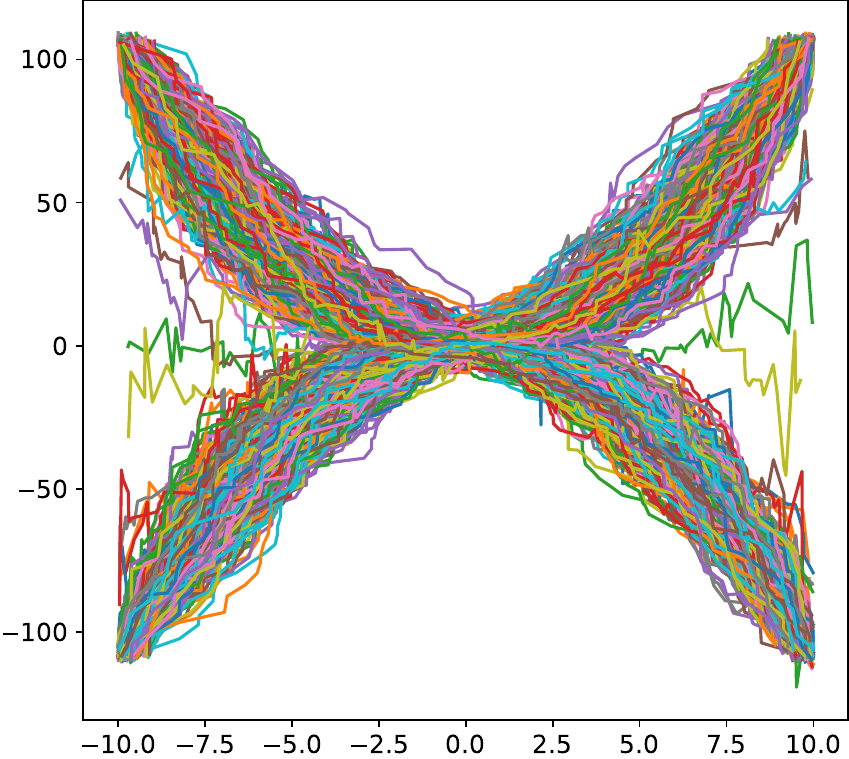}
\caption{ODE, NFE=5}
\end{subfigure}
\hfill
\begin{subfigure}{0.31\textwidth}
\centering
\includegraphics[width=\textwidth]{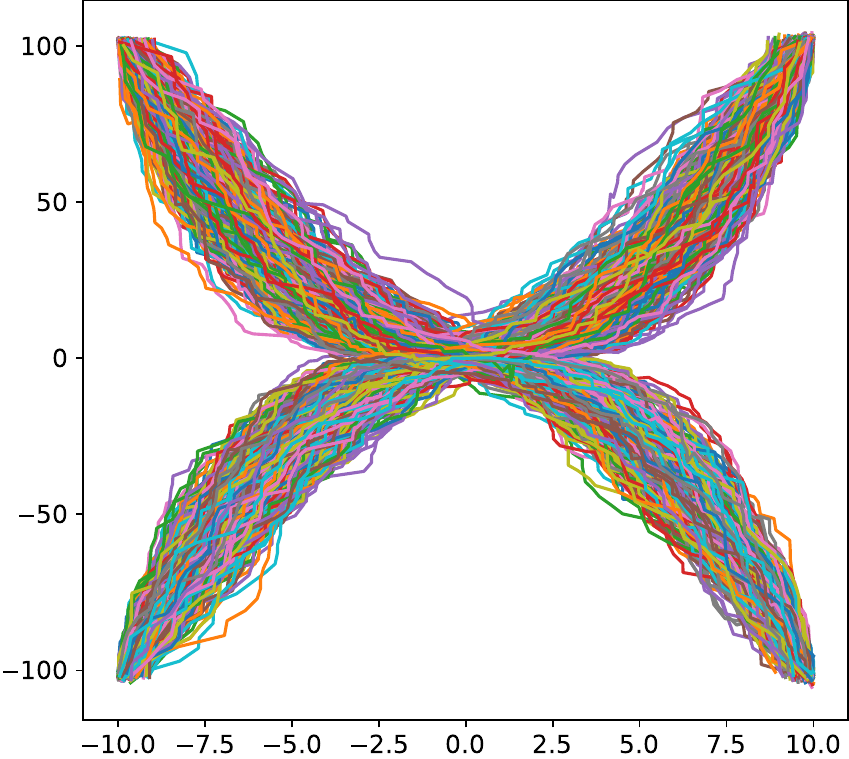}
\caption{ODE, NFE=20}
\end{subfigure}
\hfill
\begin{subfigure}{0.31\textwidth}
\centering
\includegraphics[width=\textwidth]{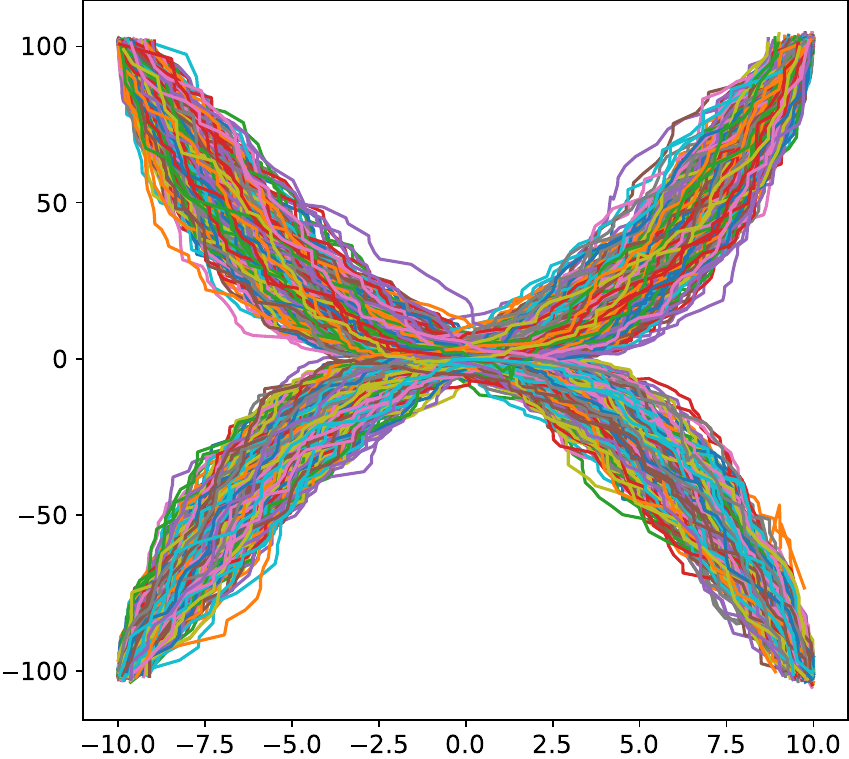}
\caption{ODE, NFE=35}
\end{subfigure}

\vspace{1em} 
\begin{subfigure}{0.31\textwidth}
\centering
\includegraphics[width=\textwidth]{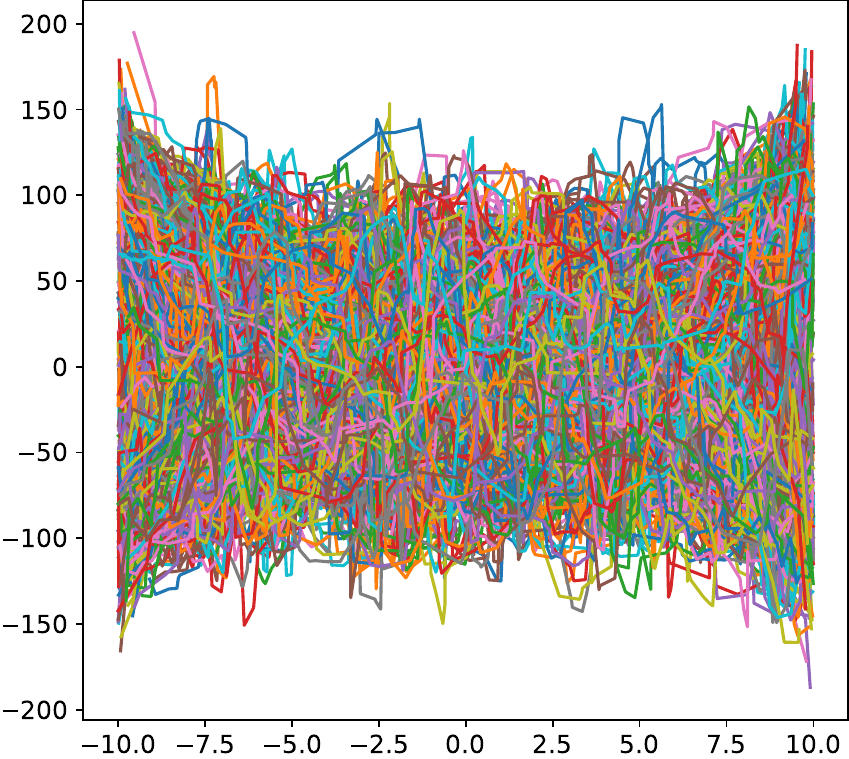}
\caption{SDE, NFE=5}
\end{subfigure}
\hfill
\begin{subfigure}{0.31\textwidth}
\centering
\includegraphics[width=\textwidth]{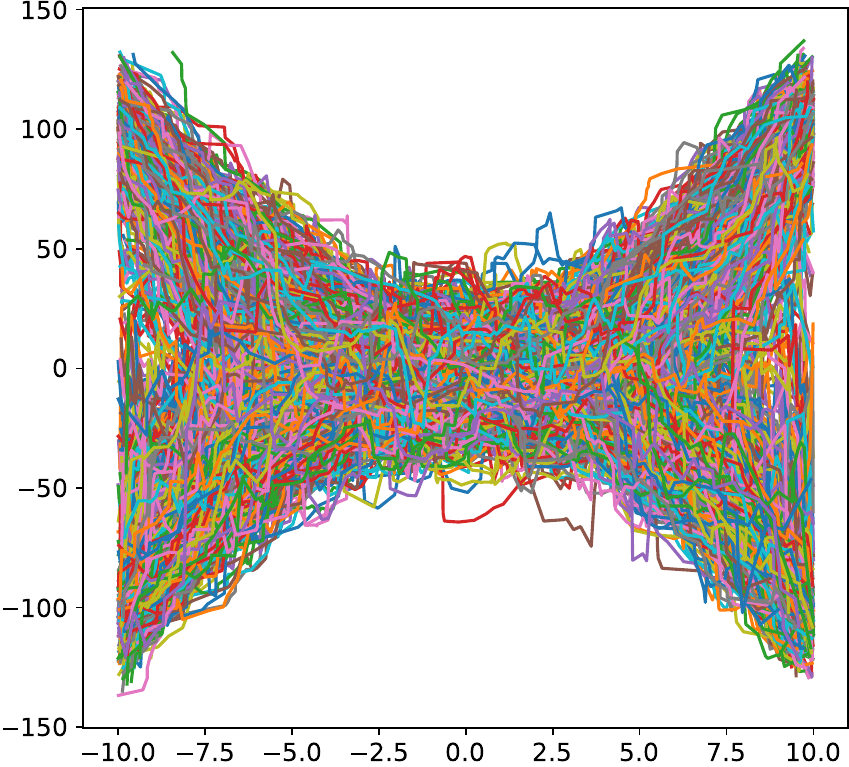}
\caption{SDE, NFE=20}
\end{subfigure}
\hfill
\begin{subfigure}{0.31\textwidth}
\centering
\includegraphics[width=\textwidth]{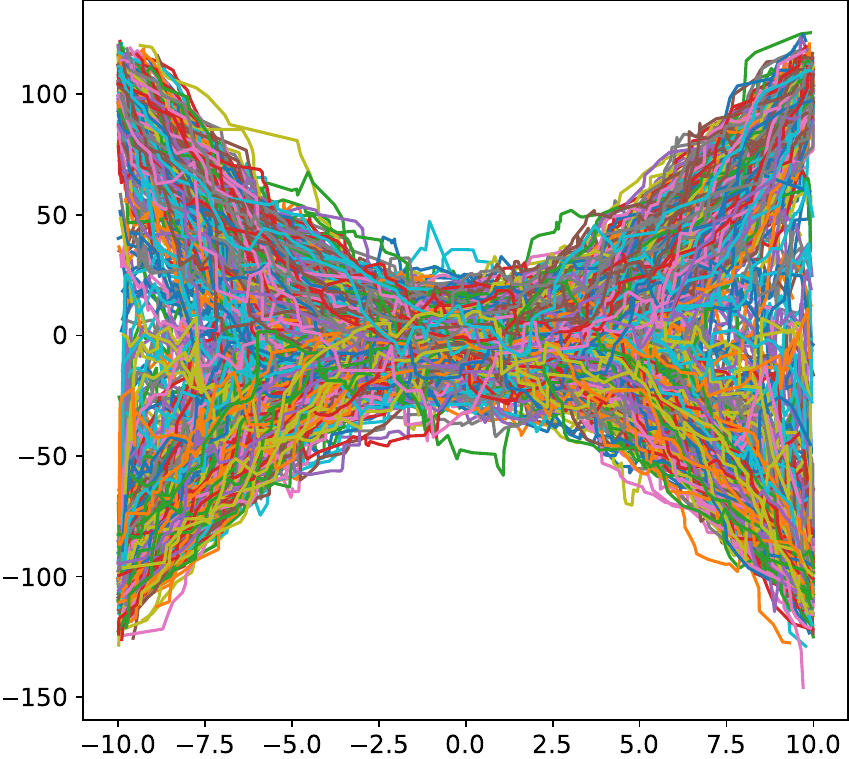}
\caption{SDE, NFE=35}
\end{subfigure}

\caption{Qualitative comparison of ODE- and SDE-generated samples in \texttt{Quadratic} dataset with  NFE$\in$\{5, 20, 35\}. Samples from the (Top) ODE solver and (Bottom) SDE solver.} 
\label{fig:Quadratic-Qualitative}
\end{figure}

\paragraph{Discussions.}
Figure~\ref{fig:Quadratic-NFE} quantitatively compares samples from the ODE and SDE solver, which clearly shows that the ODE solver outperforms the SDE solver at every considered NFE.
Remarkably, ODE solving with NFE=20 performs even better than SDE solving with \textit{any} NFE.
In Figure~\ref{fig:Quadratic-Qualitative}, we show the samples generated via the ODE and SDE solving with NFE 5, 20, and 35~(with a fixed seed). Qualitatively as well, it is clear that the ODE solver produces much better samples than the SDE solver.

\subsection{Synthetic solution of various PDEs } For the PDE tasks, we train an infinite-dimensional diffusion model via the score-matching objective. Then, we sample a synthetic solution for two well-studied PDE problems, namely, the diffusion-reaction and the heat equation, via solving our PF-ODE and the usual SDE, respectively. 
We use the same checkpoint during the inference via the ODE and SDE solving for a fair comparison. 

\subsubsection{Diffusion-reaction equation}

\paragraph{Setting.} We consider the diffusion-reaction equation of the form $$ \partial_t u = D \Delta u + R,$$ 
where $D$ is a diagonal matrix, and $R$ is a function that accounts for the diffusion of the system and the source term, respectively. 
\newpage
\begin{wrapfigure}{r}{0.48\textwidth}
  \begin{center}  
  \hspace{-4mm}
    \includegraphics[width=0.52\linewidth]{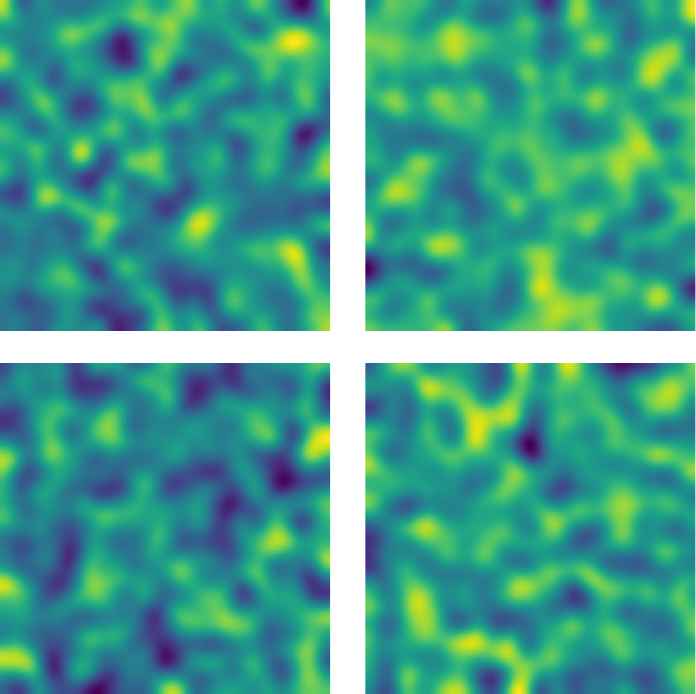}
  \end{center}
    \caption{Ground-truth solutions sampled from \texttt{PDEBench} dataset.}
    \vspace{-10mm}
    \label{fig:reacdiffGT}
\end{wrapfigure}
We utilize \texttt{PDEBench} dataset~\citep{darus-2986_2022}, which consists of solutions to the preceding diffusion-reaction equation with varying $D$ and $R$. 
\begingroup
\allowdisplaybreaks
Figure~\ref{fig:reacdiffGT} shows a batch of ground-truth solutions for diffusion-reaction equation from \texttt{PDEBench} dataset. We train an infinite-dimensional diffusion model with resolution 64. During inference, we take various NFEs $ \in \{10, 20, \cdots, 100\}$. For a quantitative investigation, we compute the sliced Wasserstein~(SW) distance~\citep{stein2024exposing} of synthetic samples~(lower SW distance is better) as in \cite{hagemann2023multilevel}. 
\endgroup

\vspace{6mm}

\begin{figure}[hbt!]
\centering

\begin{subfigure}{0.3\textwidth}
\centering
\includegraphics[width=\textwidth]{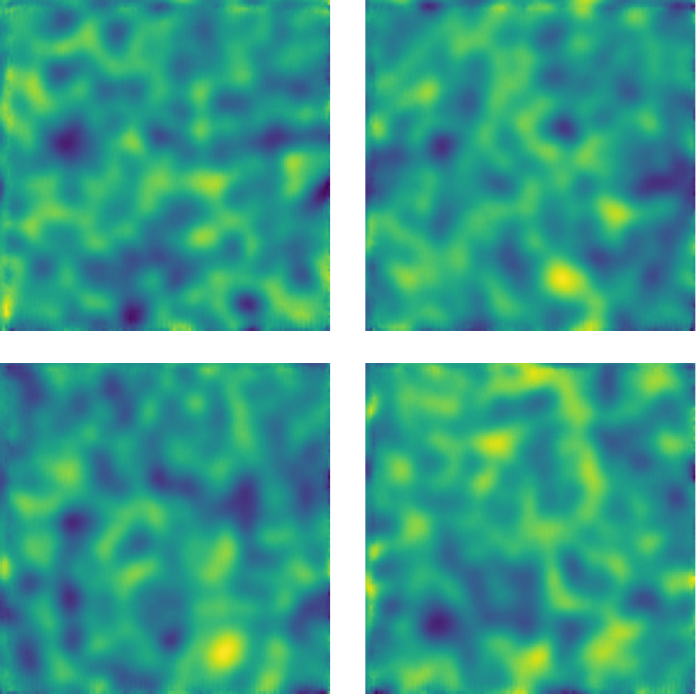}
\caption{ODE, NFE=10}
\label{fig:ODE10}
\end{subfigure}
\hfill
\begin{subfigure}{0.3\textwidth}
\centering
\includegraphics[width=\textwidth]{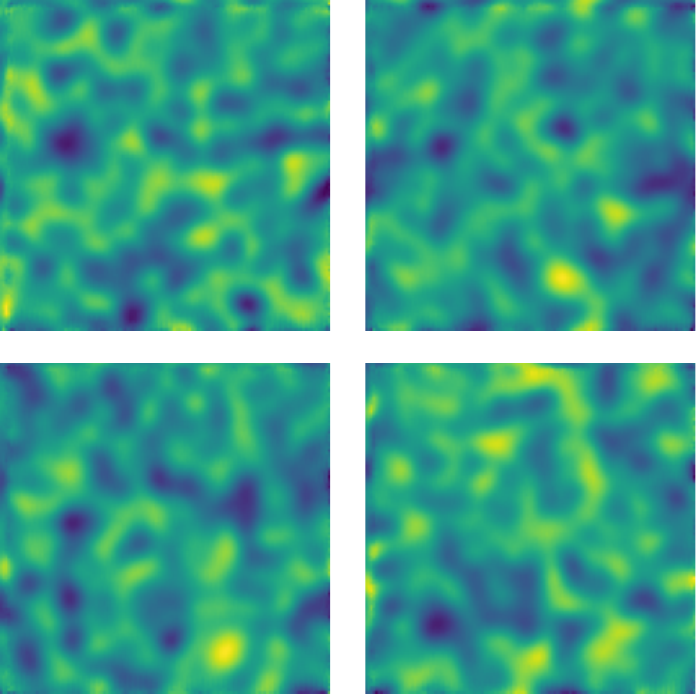}
\caption{ODE, NFE=50}
\label{fig:ODENFE50}
\end{subfigure}
\hfill
\begin{subfigure}{0.3\textwidth}
\centering
\includegraphics[width=\textwidth]{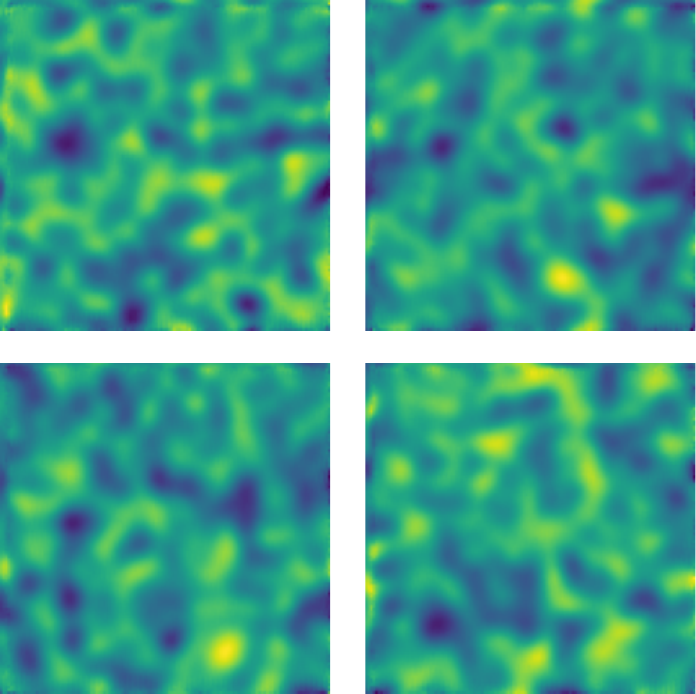}
\caption{ODE, NFE=90}
\label{fig:ODENFE90}
\end{subfigure}

\vspace{1em} 
\begin{subfigure}{0.3\textwidth}
\centering
\includegraphics[width=\textwidth]{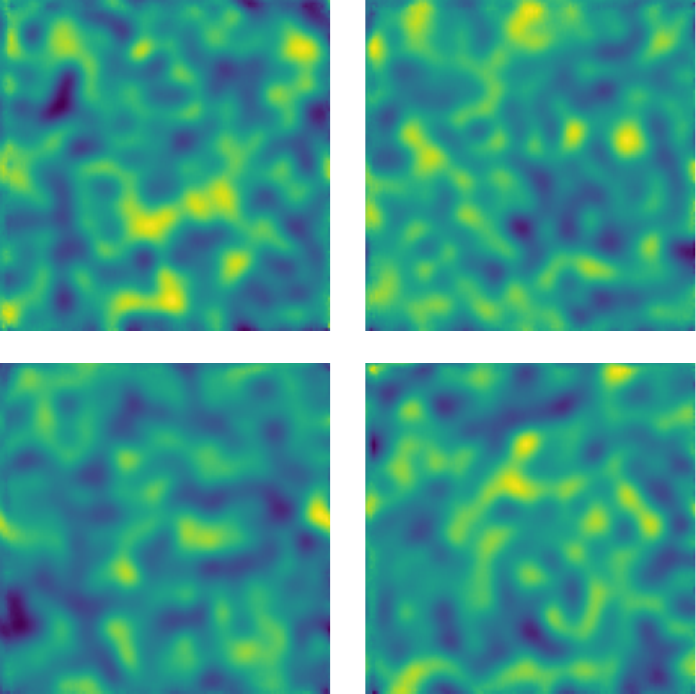}
\caption{SDE, NFE=10}
\label{fig:SDENFE10}
\end{subfigure}
\hfill
\begin{subfigure}{0.3\textwidth}
\centering
\includegraphics[width=\textwidth]{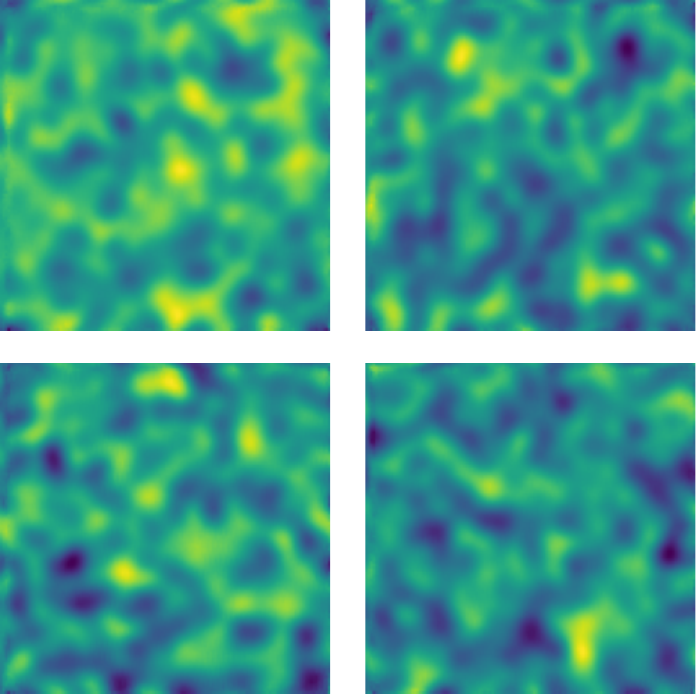}
\caption{SDE, NFE=50}
\label{fig:SDENFE50}
\end{subfigure}
\hfill
\begin{subfigure}{0.3\textwidth}
\centering
\includegraphics[width=\textwidth]{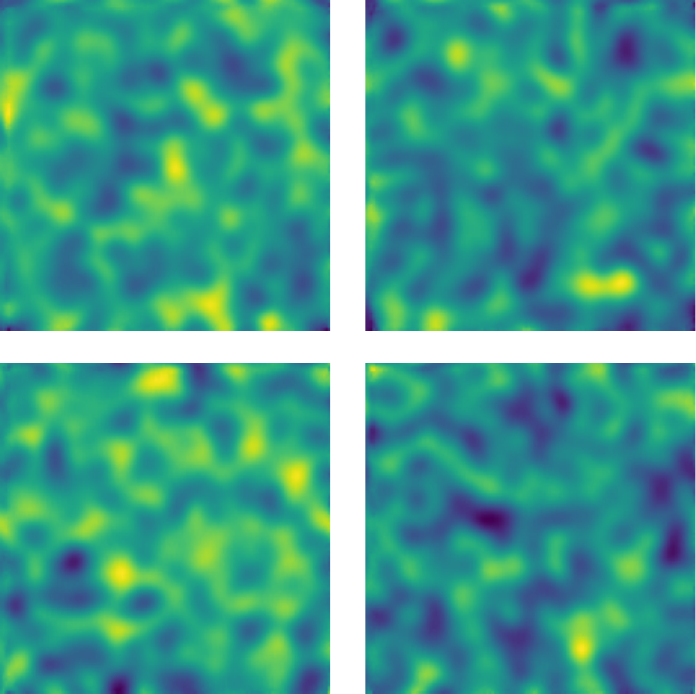}
\caption{SDE, NFE=90}
\label{fig:SDENFE90}
\end{subfigure}

\caption{Qualitative comparison of ODE- and SDE-generated solutions for diffusion-reaction equation with NFE$\in$\{10, 50, 90\}.
Samples from the (Top) ODE solver and (Bottom) SDE solver.}
\label{fig:reacdiff-Qualitative}
\end{figure}

\begin{wrapfigure}{r}{0.41\textwidth}
  \begin{center} 
  \hspace{-6mm}
\includegraphics[width=0.8\linewidth]{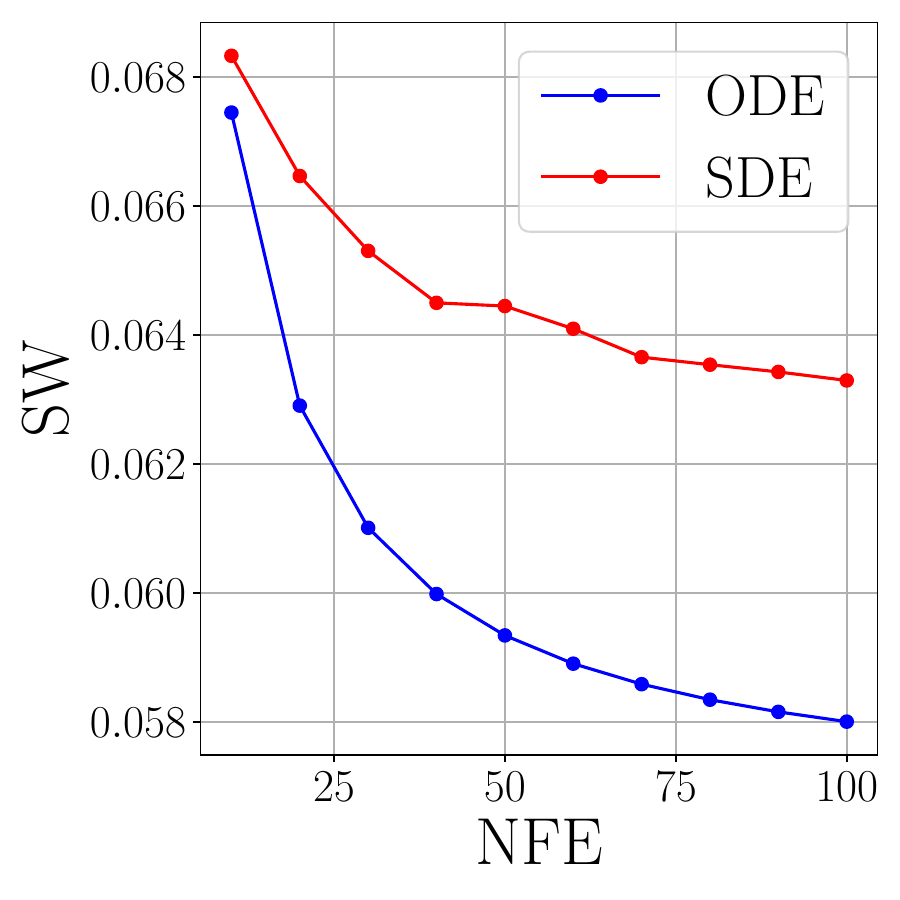}
  \end{center}
  \vspace{-4mm}
    \caption{SW vs. NFE.}
    \vspace{-20mm}
    \label{fig:reacdiff2D-quantitative}
\end{wrapfigure}

\paragraph{Discussions.} 
Figure~\ref{fig:reacdiff2D-quantitative} shows the SW distance of samples with resolution 256, generated by SDE and ODE solving at various NFEs$\in\{10, 20, \cdots, 100\}$. We note that samples from the ODE solver show a lower SW distance than those from the SDE solver at every NFE. 

Figure~\ref{fig:reacdiff-Qualitative} compares samples obtained from the ODE solver and the SDE solver at NFE 10, 50, and 90, where each samples are generated with the same fixed seed. 
Qualitatively, observe that the ODE samples across NFE $\in \{10, 50, 90\}$ (Figure~\ref{fig:ODE10},~\ref{fig:ODENFE50}, and~\ref{fig:ODENFE90}) are similar to each other, while the SDE samples across the same NFEs (Figure~\ref{fig:SDENFE10},~\ref{fig:SDENFE50}, and~\ref{fig:SDENFE90}) show severe variations.
This suggests that sampling via our PF-ODE is much faster than the SDE; for this specific example, running the ODE solver with NFE=10 is sufficient, while much more is required for the SDE solver.

\vspace{10mm}

\subsubsection{Heat equation} 

\paragraph{Setting.} We consider the heat equation on $\mathcal{O} = [-1, 1]^2$ with zero Neumann boundary condition: 
\begin{align*}
   \left\{
\begin{aligned}
    \partial_t u &= \beta \Delta u \quad \text{ on } \mathcal{O}, \\ 
    u &= 0 \quad \text{ on } \partial \mathcal{O}. 
\end{aligned}
   \right.
\end{align*}
Here, $u: [0, T] \times \mathcal{O} \to \RR$ and $\beta \in [2 \times 10^{-3}, 2 \times 10^{-2}]$ is a constant. It is well-known that the preceding heat equation, given with initial condition as an additional datum, has a unique solution under mild regularity conditions~(ref. \cite{evans2022partial}). 
Furthermore, the unique solution can be easily numerically simulated, using numerical analytic techniques such as finite difference methods~\citep{dawson1991finite}.
We generate a training dataset by randomly generating an initial condition $f $ as a mixture of sine functions (as in \cite{zhou2024masked}), and then numerically solving the heat equation with $\beta = 0.05$ and initial condition $u(0, \cdot) = f$. We train an (infinite-dimensional) diffusion model with resolution 64. For a systematic comparison, we first generate synthetic solutions $u_\mathrm{Synt}$ via the ODE or SDE solver. Then, we numerically solve the preceding heat equation via the finite-difference method with the initial condition given as $u_\mathrm{Synt}(0, \cdot)$ to obtain a ground truth solution $u_\star$. This allows us to compute the $L^p$-distance $\|u_\mathrm{Synt} - u_\star\|_{L^p([0, T] \times \mathcal{O})}$ between the synthetic solution and corresponding ground truth solution with the same initial condition. In particular, we measure their $L^2$- and $L^\infty $-distances.

\vspace{-5mm}
\begin{figure}[H]
    \centering
\includegraphics[width=\linewidth]{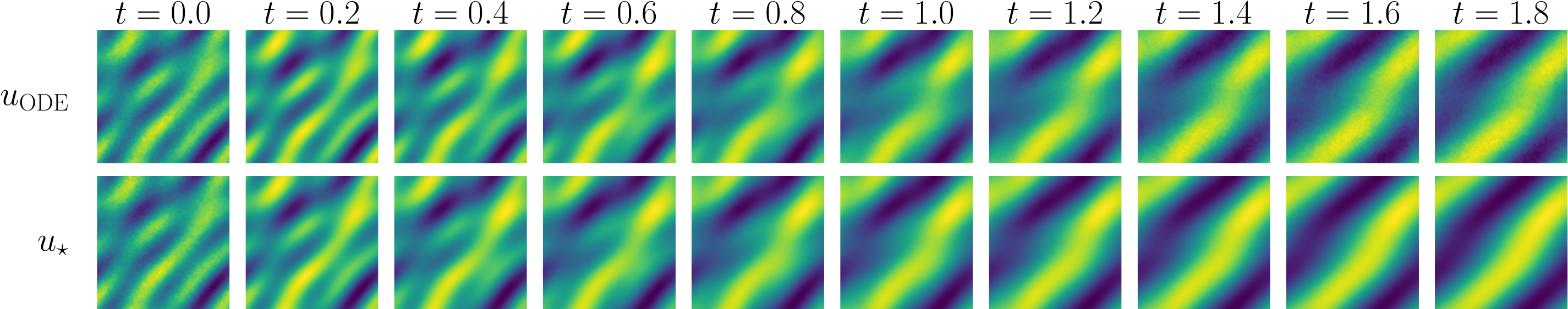}
\vspace{-5mm}
    \caption{$u_\mathrm{ODE}$ generated with NFE 10.}
    \vspace{-8mm}
    \label{fig:heat3dODE10}
\end{figure}

\vspace{-7mm}

\begin{figure}[H]
    \centering
    \includegraphics[width=\linewidth]{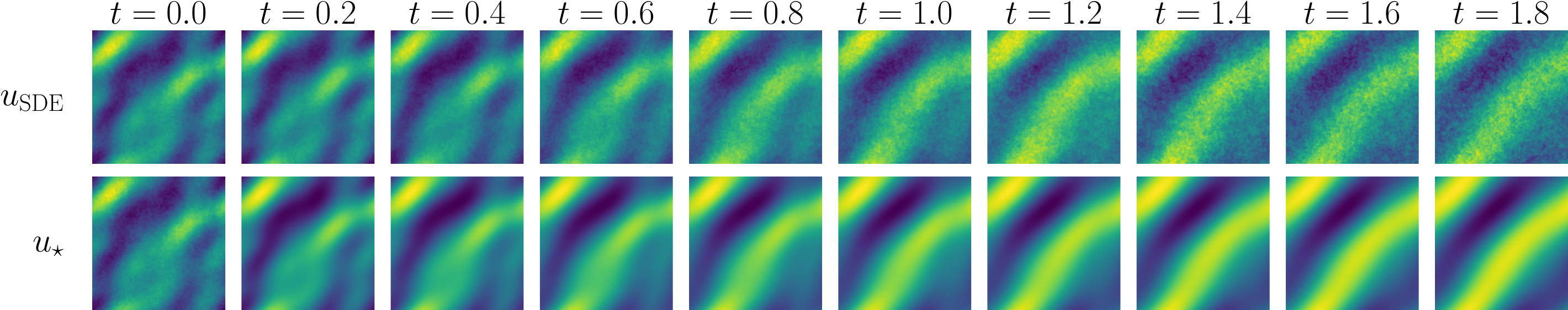}
    \vspace{-5mm}
    \caption{$u_\mathrm{SDE}$ generated with NFE 10.}
    \vspace{-8mm}
    \label{fig:heat3dSDE10}
\end{figure}

\begin{figure}[h!]
\centering
\begin{subfigure}{0.45\textwidth}
    \centering
\includegraphics[width=0.9\textwidth]{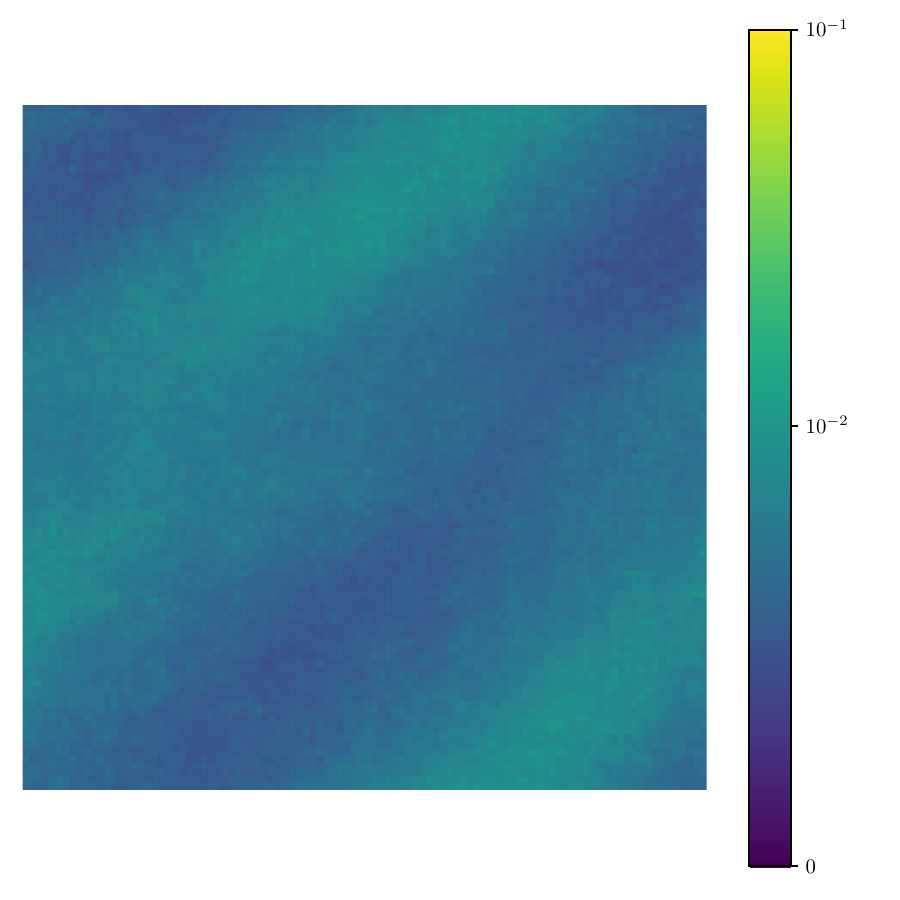}
    \caption{$|u_\mathrm{ODE}(2, \cdot) - u_\star(2, \cdot)|$}
\label{fig:heat3dQuantitative(a)}
\end{subfigure}
\hfill
\hspace{-8mm}
\begin{subfigure}{0.45\textwidth}
\centering
\includegraphics[width=0.9\textwidth]{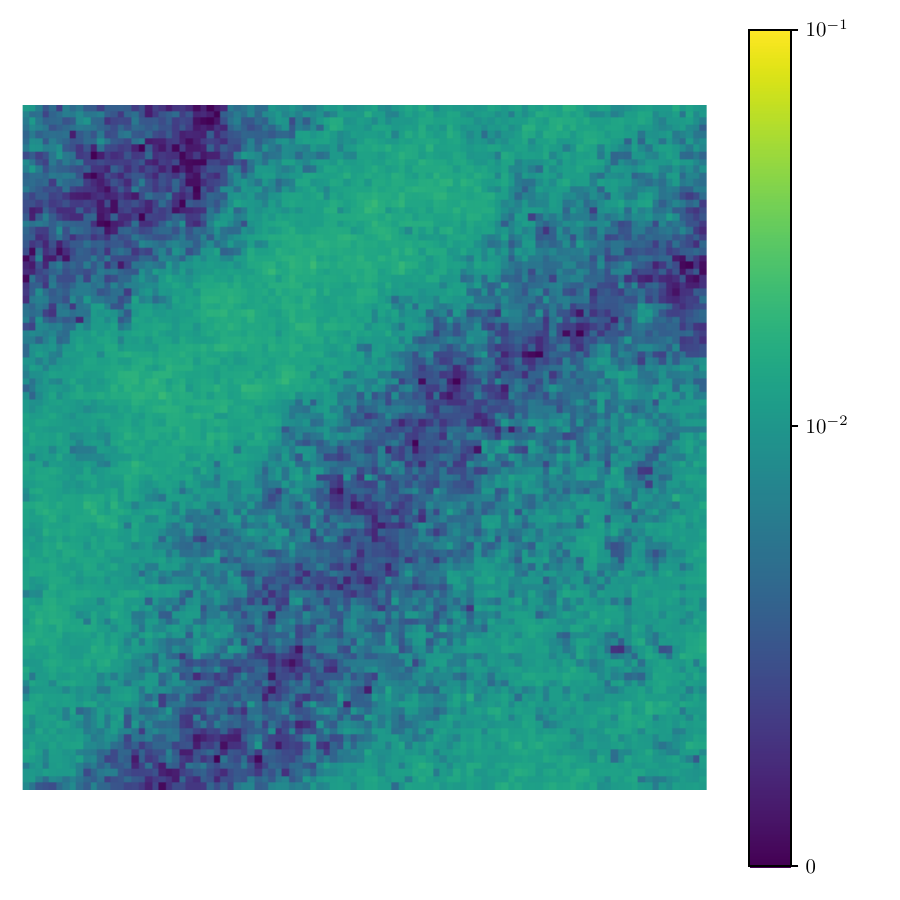}
    \caption{$|u_\mathrm{SDE}(2, \cdot) - u_\star(2, \cdot)|$}
\label{fig:Heat3dQuantitative(b)}
\end{subfigure}
\caption{Difference between a synthetic solution and the corresponding ground truth solution with same initial condition. Samples are generated with NFE 10. }
\label{fig:heat3dQuantitative}
\end{figure}

\paragraph{Discussions.} Figure~\ref{fig:heat3dODE10} shows a synthetic solution generated from the ODE solver with NFE 10 (denoted $u_\mathrm{ODE}$) and corresponding ground truth solution $u_\star$ with the same initial condition. Similarly, Figure~\ref{fig:heat3dSDE10} compares a synthetic solution obtained from solving SDE with NFE 10 and the corresponding ground truth solution. Notably, the solution generated by the ODE solver is much less noisy than that generated by the SDE solver. Figure~\ref{fig:heat3dQuantitative} shows the pixel-wise difference between a synthetic solution generated by the ODE solver and the SDE solver with the same NFE 10 and the corresponding ground truth solution with the same initial solution, where samples are generated with resolution 64. 
\vspace{-5mm}
\begin{wraptable}{r}{0.57\textwidth}
    \centering
    \begin{tabular}{@{}lcc@{}}
    \toprule
    Sampling method & ODE & SDE \\
    \midrule
    $L^2$-distance ($\downarrow$)   & \textbf{12.85}{\scriptsize ± 1.44}  & 15.58{\scriptsize ± 1.84} \\
    $L^\infty$-distance ($\downarrow$) & \textbf{1.31e-1}{\scriptsize ± 9.83e-3} & 1.46e-1{\scriptsize ± 1.36e-2} \\
    \bottomrule
    \end{tabular}
    \caption{Comparison of $L^2$- and $L^\infty$-distance for samples generated via the ODE and SDE solving with NFE 10.\vspace{-6mm}}
    \label{tab:ell-p-distance}
\end{wraptable}
\FloatBarrier
From Figure~\ref{fig:heat3dQuantitative}, one can observe that the solution generated from the ODE solver is much more similar to the ground truth than that of the SDE solver. From Table~\ref{tab:ell-p-distance}, it is notable that samples generated via the ODE solver with NFE 10 have lower $L^p$-distances to the ground truth solution $u_\star$ than those generated via the SDE solver with the same NFE.

%% file: section/04.tex
\section{Conclusion and Future Work}
In this work, we derive a notion of probability-flow ODE (PF-ODE) in infinite-dimensional function spaces with functional derivatives and measure-valued Fokker-Planck-Kolmogorov equation. 
By utilizing our infinite-dimensional PF-ODE, we lower the NFEs without affecting the sample quality in various function generation settings.
We observe that in some examples, such as time-evolving two-dimensional PDE problems, samples generated via our PF-ODE are of higher quality than those generated via the SDE not only at low NFEs but also for overall NFEs.

Our newly derived infinite-dimensional PF-ODE opens up various avenues for future work in functional diffusion models. 
First, we leave extending our work to faster sampling~\citep{lu2022dpm} and knowledge distillation~\citep{song2023consistency}) as future work.
Also, a rigorous investigation into the discretization error of infinite-dimensional diffusion models, both SDE and our PF-ODE, is another fruitful direction.
This direction may shed light on the effectiveness of our PF-ODE over SDE in several function generation tasks, which we believe is because the ODE method only incurs a single discretization error from the initial approximation of infinite-dimensional noise $\xi \sim \mathcal{N}(0, Q)$; in contrast, for the SDE method, repeated discretization error for $\xi$ occurs.

%% file: section/A01.tex
\newpage

\begin{center}
    \Large{\bf \textsc{Appendix}} 
\end{center}

\tableofcontents

\normalsize 
\newpage
\section*{Table of notations}

\begin{table}[h]
    \centering
    \begin{tabular}{p{2.5cm} p{10cm}}
        \toprule
        Symbol & Description \\
        \midrule
        $T_\# \mu$ & Pushforward of a measure $\mu$ by a map $T$ \\
        $X^\ast$ & Dual space of $X$ \\
        $\langle x^\ast, x\rangle$ & Dual pairing of $x^\ast \in X^\ast$ and $x \in X$ \\
        $\hat{\mu}$ & Characteristic function of a probability measure $\mu$ on $X$ defined by $\hat{\mu}(h) = \int e^{i \langle h, x \rangle} \mu(dx)$, $h \in X^\ast.$ \\
        $\langle \cdot, \cdot \rangle_{\mathcal{H}}$ & Inner product on a Hilbert space $\mathcal{H}$ \\
        $\operatorname{Tr}_{\mathcal{H}}$ & Trace on  a Hilbert space $\mathcal{H}$ \\ 
        $\mathscr{L}_2(\mathcal{H})$ & The set of Hilbert-Schmidt operators on $\mathcal{H}$ \\
        $\mathcal{N}(0, Q)$ & Centered Gaussian measure in $\mathcal{H}$ with covariance operator $Q$ \\
        $(W_t)_{t \ge 0}$ & A $Q$-Wiener process in $\mathcal{H}$ \\
        $\mathcal{H}_Q$ & The Cameron-Martin space of $\mathcal{N}(0, Q)$ \\
        $\mathcal{FC}_b^\infty(\mathcal{H})$ & The set of all cylindrical functions on $\mathcal{H}$ \\
        $\gM(\gH)$ & The set of all Borel measures on $\mathcal{H}$ \\
        $\mathcal{L}_t$ & Kolmogorov operator defined on $\mathcal{FC}_b^\infty(\mathcal{H})$ \\
        $\partial_h f(x)$ & G\^ateaux differential of $f$ at $x$ along $h$ \\ 
        $Df(x)$ & Fr\'echet derivative of $f$ at $x$ \\ 
        $\mathrm{Law}(X)$ & Distribution~(law) of a random variable $X$ \\
        $\rho_{\mathcal{H}_Q}^{\mu}$ & Logarithmic gradient of $\mu$ along $\mathcal{H}_Q$ \\
        $\Delta$ & The Laplace operator \\
        $\nabla$ & The gradient operator \\
        \bottomrule
    \end{tabular}
    \caption{Mathematical Symbols and Definitions}
    \label{tab:symbols}
\end{table}

%
\newpage
\section{Mathematical Preliminaries}
\label{app:prelim}

In this section, we provide a gentle introduction to the theory of Gaussian measures and stochastic processes in infinite dimensional Hilbert spaces. Most of the content of this section can be found in \cite{da2014stochastic}, \cite{bogachev1998gaussian}, \cite{bogachev2022fokker}, \cite{prevot2007concise}, or \cite{kuo1975gaussian}.

We introduce several notations and definitions here before introducing precise definitions of mathematical objects we exploit in this research. 

\paragraph{Pushforward measure.} If $(X, \mathcal{F})$ and $(Y, \mathcal{G})$ are measurable spaces and $T : X \to Y$ is $\mathcal{F}/\mathcal{G}$-measurable, then for any measure $\mu$ on $(X, \mathcal{F})$ we define the pushforward measure $T_\# \mu $ by
\begin{align*}
    (T_\# \mu)(A) = \mu (T^{-1}(A)), \quad \forall A \in \mathcal{G}. 
\end{align*}

\paragraph{Duality and pairing.} For a locally convex topological vector space $X$ over $\Bbbk = \RR \text{ (or } \mathbf{C})$, we denote by $X^\ast$ the dual space of $X$, i.e., 
\begin{align*}
    X^\ast = \{ \ell : X \to \Bbbk \mid \ell \text{ is linear and continuous} \}. 
\end{align*}
For $\ell \in X^\ast$ and $x \in X$ we denote by $\langle \ell, x\rangle$ the quantity $\ell(x)$.

\paragraph{Characteristic function.} 
If $\mu$ is a probability measure on $(X, \mathcal{B}(X))$, we define the characteristic function $\hat{\mu}$ of $\mu$ by 
\begin{align*}
    \hat{\mu}(h) = \int_{X} e^{i \langle h, x \rangle}  \mu(dx), \quad \forall h \in X^\ast.  
\end{align*}
It is well known that if $\hat{\mu} = \hat{\nu},$ then $\mu = \nu$~(ref. \cite{da2014stochastic}, Proposition 2.5).
\vspace{5mm}

\subsection{Gaussian measures and Wiener processes}

\subsubsection{Gaussian measures}

\begin{definition}
 A Borel probability measure $\mu$ on a locally convex space $X$ is called a Gaussian measure if the pushforward measure $h_\# \mu$ is Gaussian for every $h \in X^\ast$. The measure $\mu$ is said to be centered if $h_\# \mu$ is centered in $\RR$ for every $h \in X^\ast$. 
\end{definition}

\begin{theorem}[ref. \cite{bogachev1998gaussian}, Theorem 2.2.4] A measure $\mu$ on a locally convex space $X$ is Gaussian if and only if its characteristic function is of the form 
\begin{align*}
    \hat{\mu}(h) = \exp\left[ 
    i L(h) - \frac{1}{2}B(h, h)
    \right], \quad \forall h \in X^\ast, 
\end{align*}
where $L$ is a linear functional on $X^\ast$ and $B$ is a symmetric, non-negative bilinear form on $X^\ast$. 
\end{theorem}

From now on, we stick to the case where $X = \mathcal{H}$ is a separable Hilbert space. In this case, we may identify $\mathcal{H}$ with $\mathcal{H}^\ast$ via the Riesz representation. If $\mu$ is a Gaussian measure on $\mathcal{H}$, we can find some $m\in \mathcal{H}$ and a non-negative symmetric operator $Q : \mathcal{H} \to \mathcal{H}$ such that 
\begin{align*}
    \hat{\mu}(h) = \exp\left[
    i \langle m, h \rangle_\mathcal{H} - \frac{1}{2} \langle Qh, h \rangle_\mathcal{H} 
    \right] \quad \forall h \in \mathcal{H}. 
\end{align*}
It is known that for every $f, g \in\mathcal{H}$, 
\begin{align*}
    \langle m, f \rangle_\mathcal{H}& = \int_\mathcal{H} \langle f, x \rangle_\mathcal{H} \mu(dx)  \\
    \langle Qf, g \rangle_\mathcal{H} &= \int_{\mathcal{H}} \langle f, x - m \rangle_\mathcal{H} \langle g, x - m \rangle_\mathcal{H} \mu(dx)
\end{align*}
In other words, if $Z$ is an $\mathcal{H}$-valued random variable with $\operatorname{Law}(Z) = \mu$, then 
\begin{align*}
    m = \mathbb{E}[Z], \quad Q = \operatorname{Cov}(Z).
\end{align*}
We call $m$ the mean vector and $Q$ the covariance operator, and write $\mu = \mathcal{N}(m, Q)$.

We end this subsection by providing a brief notes on Cameron-Martin space of a Gaussian measure. Although there are several equivalent definitions of Cameron-Martin space, we follow that of \cite{da2014stochastic} as it can be presented without providing additional technical details. 

\begin{definition}
    Let $\mu$ be a centered Gaussian measure on a locally convex space $X$. A linear space $\mathcal{H}_\mu \subset X$ equipped with an inner product is called a Cameron-Martin space of $\mu$ if $\mathcal{H}_\mu$ is continuously embedded in $X$ and for every $h \in X^\ast$, one has that $\operatorname{Law}(\varphi) = \mathcal{N}(0, |\varphi|_\mu^2)$, where 
    \begin{align*}
        |\varphi|_\mu = \sup_{h \in \mathcal{H}_\mu, \|h\|_{\mathcal{H}_\mu} \le 1} |\varphi(h)|. 
    \end{align*}
\end{definition}

For a Gaussian measure $\mu = \mathcal{N}(0, Q)$, it is known that the Cameron-Martin space $\mathcal{H}_\mu = \mathcal{H}_{\mathcal{N}(0, Q)}$ is given by 
\begin{align*}
    \mathcal{H}_\mu = Q^{1/2}(\mathcal{H}), \quad \langle f, g \rangle_{\mathcal{H}_\mu} = \langle Q^{-1/2}f, Q^{-1/2} g \rangle_\mathcal{H}. 
\end{align*}
We shall simply denote $\mathcal{H}_\mu =\mathcal{H}_Q$ in this case. If $\{\varphi_i\}$ is an orthonormal basis of the ambient Hilbert space $\mathcal{H}$, then $\{Q^{1/2} \varphi_i\}$ becomes an orthonormal basis for the Cameron-Martin space $\mathcal{H}_Q$.

\vspace{3mm}

\subsubsection{Wiener processes}

\begin{definition}
    Let $Q$ be a trace class non-negative symmetric operator on $\gH$. An $\gH$-valued stochastic process $W = (W_t)_{t \in [0, T]}$ on a probability space $(\Omega, \mathcal{F}, \PP)$ is called a standard $Q$-Wiener process, if 
    \begin{itemize}
        \item[1.] $W(0) = 0$, 
        \item[2.] $W$ has continuous trajectories, i.e., $W$ has $\PP$-continuous paths, 
        \item[3.] $W$ has independent increments, i.e., for any $n \in \NN$ and $0 < t_1 < \cdots < t_n < \infty$, 
        \begin{align*}
            W_{t_1}, \ W_{t_2} - W_{t_1}, \cdots, \quad W_{t_n} - W_{t_{n - 1}} 
        \end{align*}
        are independent, 
        \item[4.] the increments have the following Gaussian laws: 
        \begin{align*}
            \mathbb{P} \circ (W_{t} - W_s)^{-1} := \operatorname{Law}(W_t - W_s) = \mathcal{N}(0, (t - s)Q) 
        \end{align*}
        for all $0 \le s \le t \le T$. 
    \end{itemize}
\end{definition}

In this work, $Q$ always denote a trace class non-negative symmetric operator on $\gH$. Notice that $Q$ is diagonalizable, and in particular, there exists a sequence $\{\varphi_k\}_{k = 1}^\infty $ consists of eigenvectors of $Q$ and a sequence of non-negative real numbers $\{\lambda_k\}_{k = 1}^\infty$ such that $Q\varphi_k = \lambda_k \varphi_k$ for all $k = 1, 2, \cdots$~\citep[Chapter II]{conway2019course}. Based on this eigensystem of $Q$, one can express $Q$-Wiener process as a series expansion. More precisely, one has the so-called Kosambi–Karhunen–Loève Theorem (see \cite{prevot2007concise}, for example): 

\begin{theorem}
    A $\gH$-valued stochastic process $W = (W_t)_{t \ge 0}$ is a $Q$-Wiener process if and only if 
    \begin{align*}
        W_t = \sum_{k = 1}^\infty \sqrt{\lambda_k} \beta^k_t \varphi_k 
    \end{align*}
    where $\beta^k = (\beta^k_t)_{t \in [0, T]}$ are independent real-valued Brownian motions on a probability space $(\Omega, \mathcal{F}, \PP)$. The series converges in $L^2(\Omega, \mathcal{F}, \PP; C([0, T]; \mathcal{U}))$. (Hence, there exists a $\PP$-a.s. continuous version of $W$.) 
\end{theorem}

\vspace{5mm}

\subsection{Functional Derivatives and Fomin derivative}

In this subsection, we introduce the notion of Fr\'echet and G\^ateaux derivative (of functions $\mathcal{H} \to \RR$) and Fomin derivative (of Borel measures on $\mathcal{H}$). Contents of this section can be found in \cite{helin2015maximum} and \citet[Chapter 10]{bogachev2022fokker}, for example.

\subsubsection{Functional derivatives}

Let $X$ and $Y$ be locally convex spaces, and let $U \subset X$ be open. For a function $F: U \to Y$, the G\^ateaux differential of $F$ along $h \in X$ is defined by 
\begin{align*}
    \partial_h F(u) := \lim_{\varepsilon \to 0} \frac{F(u + \varepsilon h) - F(u)}{\varepsilon} = \frac{d}{d\varepsilon}\bigg|_{\varepsilon = 0} F(u + \varepsilon h), 
\end{align*}
whenever the limit exists. If the limit exists for every $h \in X$, then $F$ is called G\^ateaux differentiable at $u$.

In this paper, we stick to the cases where $X = \mathcal{H}$ and $Y = \RR$ or $Y = \mathcal{H}$. 
In these cases~(or more generally whenever $X$ and $Y$ are normed spaces), there is another canonical notion of differentiability called the Fr\'echet derivative. For a function $F : X \to Y$ (where $X$ and $Y$ are normed spaces), we say that $F$ is Fr\'echet differentiable at $x \in U$ if there is a bounded linear operator $DF(x): X \to Y$ such that 
\begin{align*}
    \lim_{\|h\| \to 0} \frac{\|F(x + h) - F(x) -  DF(x)h\|}{\|h\|} = 0. 
\end{align*}
The notion of Fr\'echet differentiability is stronger than that of G\^ateaux differentiability in the sense that whenever $F$ is Fr\'echet differentiable at $x \in X$, then $F$ is G\^ateaux differentiable at $x$ too, and $\partial_h F(x) = DF(x)(h)$. In particular, when $X = \mathcal{H}$ and $Y = \RR$, then we can equivalently understand the notion of Fr\'echet differentiability at $x \in \mathcal{H}$ as an existence of $DF(x) \in \mathcal{H}$ such that
\begin{align*}
    \lim_{\|h\| \to 0} \frac{|F(x + h) -F(x) - \langle DF(x), h\rangle_{\mathcal{H}}|}{\|h\|} = 0, 
\end{align*}
via the Riesz isomorphism $\mathcal{H} \cong \mathcal{H}^\ast$. 
If $F : \mathcal{H} \to \RR$ is Fr\'echet differentiable at $x$, then its G\^ateaux differential along $h \in \mathcal{H}$ coincides with $\langle DF(x), h\rangle$.

\subsubsection{Fomin derivative}
In this subsection, we briefly introduce the notion of Fomin differentiability of measures, which is developed by \cite{fomin1968differentiable}. 
Although one can define the notion of Fomin differentiability for any Borel probability measure $\mu$ on a locally convex space $X$, we will stick to the case where $X = \mathcal{H}$ (and $\mu \in \mathcal{M}(\mathcal{H})$).

In an infinite dimensional space $\mathcal{H}$, neither the natural notion of probability density function ($p$) nor the notion of gradient ($\nabla$) exists. Still, a notion of the logarithmic gradient of probability measure $\mu$ on $\mathcal{H}$ (that acts as $\nabla \log p$) exists. We provide the formal definition below. 

\begin{definition}[ref. \cite{bogachev1999absolutely}]
    Let $\mu$ be a (Borel) probability measure on $\mathcal{H}$, and let $\mathcal{K} \subset \mathcal{H}$ be a densely embedded Hilbert space. We say $\mu$ is Fomin differentiable along $h \in \mathcal{K}$ if there exists a function $\rho_h^\mu \in L^1(\mu)$ such that 
    \begin{align}
        \int_\mathcal{H} \partial_h f_{\varphi_1, \cdots, \varphi_n}(x) \mu(dx) = -\int_\mathcal{H} f_{\varphi_1, \cdots, \varphi_n}(x)  \rho_h^\mu (x) \mu(dx). 
        \label{eq:ibp}
    \end{align}
    If there exists a function $\rho_{\mathcal{K}}^\mu : \mathcal{H} \to \mathcal{H}$ such that $\langle \rho_{\mathcal{K}}^\mu(x), h \rangle_{\mathcal{K}} = \rho_h^{\mu}(x)$
    for every $x \in \mathcal{H}$ and $h \in \mathcal{K}$, then we call $\rho_\mathcal{K}^\mu$ the logarithmic gradient of $\mu$ along $\mathcal{K}$. 
\end{definition}

It is well-known that $\mu$ is differentiable along $h$ in the sense of Fomin if and only if the following quantity 
\begin{align*}
    d_h \mu(A) = \lim_{\varepsilon \to 0} \frac{\mu(A + \varepsilon h) - \mu(A)}{\varepsilon} 
\end{align*}
exists for every Borel set $A$ in $\mathcal{H}$~(ref. \cite{helin2015maximum}, Proposition 1). Because the zero measure is the only measure on $\mathcal{H}$ which is Fomin differentiable along every vector in $\mathcal{H}$~\citep[p.406]{bogachev2022fokker}, it is a necessary treatment in the above \eqref{eq:ibp} to specify the set $\mathcal{K} \subset \mathcal{H}$.

%% file: section/A02.tex
\newpage

\section{Proof of Theorem~\ref{thm:prob-flow-ode}}
\label{app:proof-pfode}

This section provides a rigorous proof of our main theoretical result, \thmref{thm:prob-flow-ode}, which we re-state for the sake of convenience. 

\begin{theorem*}[Restatement of \thmref{thm:prob-flow-ode}]
    Let $X_t$ be a solution of an SDE in $\mathcal{H}$ of the form 
    \begin{align*}
        dX_t = B(t, X_t) dt + G(t) dW_t, \quad X_0 \sim \PP_0 = \PP_\mathrm{data}
    \end{align*}
    and let $\mu_t := \mathrm{Law}(X_t)$.
    Then, $\mu_t$ satisfies the Fokker-Planck-Kolmogorov equation of $(Y_t)_{t \in [0, T]}$, where $(Y_t)_{t \in [0, T]}$ is a solution of the following {(infinite dimensional) probability-flow ODE} 
    \begin{align}
       dY_t = \left[ 
       B(t, Y_t)-  \frac{1}{2} A(t) \rho_{\mathcal{H}_Q}^{\mu_t}(Y_t)
       \right] dt, \quad Y_0 \sim \PP_0.
       \label{eq:1-1}
    \end{align}
    Here, $A(t) := G(t) G(t)^\ast$ and $\rho_{\mathcal{H}_Q}^{\mu_t}$ is the logarithmic gradient of $\mu_t$ along the Cameron-Martin space $\mathcal{H}_Q$ of $\mathcal{N}(0, Q)$. 
\end{theorem*}

In the proof of \thmref{thm:prob-flow-ode}, we slightly abuse notations and write $\varphi(h) = \langle \varphi, h \rangle$ in order to avoid too many brackets in the presentation. Also, we utilize the dual pairing notation of a function and a Borel (probability) measure: We write $\langle f, \nu\rangle$ for $f \in \mathcal{FC}_b^\infty(\mathcal{H})$ and $\nu \in \mathcal{M}(\mathcal{H})$ to denote the quantity $\int_{\mathcal{H}} f(u) \nu(du)$.

\begin{proof}[Proof of \thmref{thm:prob-flow-ode}]
View the probability-flow ODE (\eqref{eq:1-1}) as an SDE in $\mathcal{H}$ with no diffusion term, and consider the associated Kolmogorov type operator $\Tilde{\mathcal{L}}_t$, $t \in [0, T]$, defined as 
\begin{align*}
       \Tilde{\mathcal{L}}_t f_{\varphi_{1}, \cdots, \varphi_m}(u) = \bigg<
       D f_{\varphi_{1}, \cdots, \varphi_m} (u), \underbrace{B(t, u) - \frac{1}{2} A(t) \rho_{\mathcal{H}_Q}^{\mu_t} (u) }_{=: \mathcal{G}(t, u)}
      \bigg>_{\mathcal{H}_Q}, \quad \forall f_{\varphi_{1}, \cdots, \varphi_m} \in \mathcal{FC}_b^\infty(\mathcal{H}), 
\end{align*}
where $Df_{\varphi_{1}, \cdots, \varphi_m}$ stands for the Fr\'echet derivative of $f_{\varphi_{1}, \cdots, \varphi_m}$. 
 One has to check if the time-evolution of $\mu_t = \mathrm{Law}(X_t)$ can be described in terms of the Kolmogorov type operator $\Tilde{\mathcal{L}}_t$. That is, one has to check if $\mu_t$ satisfies 
\begin{align}
    \partial_t \langle f_{\varphi_{1}, \cdots, \varphi_m}, \mu_t\rangle =\langle \Tilde{\mathcal{L}}_t f_{\varphi_{1}, \cdots, \varphi_m}, \mu_t\rangle, \quad \forall f_{\varphi_{1}, \cdots, \varphi_m}\in \mathcal{FC}_b^\infty(\mathcal{H}). 
    \label{eq:check}
\end{align}
By the definition of the Kolmogorov operator $\Tilde{\mathcal{L}}_t$, one expands the left-hand side of \eqref{eq:check} as 
\begin{align*}
    &\left\langle \Tilde{\mathcal{L}}_t f_{\varphi_{1}, \cdots, \varphi_m}, {\mu}_t\right\rangle\\
     =& \int_{\mathcal{H}} \left< 
    D f_{\varphi_{1}, \cdots, \varphi_m} (u), \mathcal{G}(t, u) 
    \right>_{\mathcal{H}_Q}\mu_t(du) \\
    =& \underbrace{\int_{\mathcal{H}} \left[\left<
    Df_{\varphi_{1}, \cdots, \varphi_m} (u), B(t, u)
    \right>_{\mathcal{H}_Q} - \frac{1}{2} \left<
    Df_{\varphi_{1}, \cdots, \varphi_m} (u), A(t)\rho_{\mathcal{H}_Q}^{\mu_t} (u) 
    \right>_{\mathcal{H}_Q} \right] \mu_t(du)}_{=:(\mathrm{I})}.
\end{align*}
On the other hand, note that one already knows that $\{\mu_t\}_{t \in [0, T]}$ satisfies the Kolmogorov forward equation for the original stochastic differential equation. That is, if we define 
    \begin{align*}
        \mathcal{L}_t f_{\varphi_{1}, \cdots, \varphi_m}(u) = \frac{1}{2} \operatorname{Tr}_{\mathcal{H}_Q} \left(
        A(t) \circ Q \circ D^2 f_{\varphi_{1}, \cdots, \varphi_m}(u)
        \right) + \left<
        D f_{\varphi_{1}, \cdots, \varphi_m}(u), B(t, u) \right>_{\mathcal{H}_Q} 
    \end{align*}
    for $f_{\varphi_{1}, \cdots, \varphi_m} \in \mathcal{FC}_b^\infty(\mathcal{H})$, 
then the time-evolution of $\{\mu_t\}_{t \in [0, T]}$ can be expressed as the following Cauchy problem in a weak sense~\citep{belopolskaya2012stochastic}: 
    \begin{align}
        \left\{ 
        \begin{aligned}
             &\partial_t {\mu}_t = (\mathcal{L}_t)^\ast {\mu}_t, \quad t \in (0, T), \\ 
             & {\mu}_t\Big|_{t = 0 } = \PP_0, 
        \end{aligned}
      \right.
      \label{eq:forward}
    \end{align}
    where $(\mathcal{L}_t)^\ast$ denotes the formal adjoint of $\mathcal{L}_t$. From the forward equation (\eqref{eq:forward}), one has that 
    \begin{align*}
         &\partial_t \left< f_{\varphi_{1}, \cdots, \varphi_m}, \mu_t\right>\\ =&\underbrace{ \int_{\mathcal{H}} \left[ 
       \frac{1}{2}\operatorname{Tr}_{\mathcal{H}_Q} \left(
        A(t) \circ Q \circ D^2 f_{\varphi_{1}, \cdots, \varphi_m}(u)
        \right) + \left< 
        D f_{\varphi_{1}, \cdots, \varphi_m}(u), B(t, u) \right>_{\mathcal{H}_Q}
         \right]\mu_t(du)}_{=: (\mathrm{II})}
    \end{align*}
    for every $f_{\varphi_{1}, \cdots, \varphi_m} \in \mathcal{FC}_b^\infty(\mathcal{H})$. 
Hence, in order to prove \eqref{eq:check}, one has to check if $(\mathrm{I}) = (\mathrm{II})$. To establish this result, it suffices to prove the following 

\textbf{Claim.} \textit{It holds that  
    \begin{align}
        -\int_{\mathcal{H}}\left< 
    D f_{\varphi_{1}, \cdots, \varphi_m}(u) , A(t) \rho_{\mathcal{H}_Q}^{\mu_t} (u)
    \right>_{\mathcal{H}_Q} \mu_t(du) 
    \overset{(!)}{=} \int_\mathcal{H}  \operatorname{Tr}_{\mathcal{H}_Q} \left(
    A(t) \circ Q \circ D^2 f_{\varphi_{1}, \cdots, \varphi_m}(u)
    \right) \mu_t(du). 
    \label{eq:claim}
\end{align}} 

\vspace{2mm}
Before proving the preceding claim, we first establish an auxiliary result on the first- and second-order Fr\'echet derivatives of cylindrical functions. We defer the proof of the following Lemma to the end of this section. 
\begin{lemma}
    For $f_{\varphi_1, \cdots, \varphi_m}  \in \mathcal{FC}_b^\infty(\mathcal{H})$, $f \in \mathcal{C}_0^\infty(\RR^m)$, its first- and second-order Fr\'echet derivatives are given by 
    \begin{align*}
         Df_{\varphi_1, \cdots, \varphi_m}(u) &= \sum_{i = 1}^m (\partial_i f)(\langle \varphi_1, u \rangle, \cdots, \langle \varphi_m, u \rangle) \varphi_i, \\
          D^2f_{\varphi_1, \cdots, \varphi_m}(u) (h) &= \sum_{i, j = 1}^m \varphi_i  (\partial_{ij}^2 f)(\langle \varphi_1, u\rangle, \cdots, \langle \varphi_m, u \rangle) \langle \varphi_j, h \rangle, \quad h \in \mathcal{H}. 
    \end{align*}

    \label{Lemma}
\end{lemma}

We then have
\begingroup
\allowdisplaybreaks
\begin{align*}
    &\operatorname{Tr}_{\mathcal{H}_Q} \left(
    A(t) \circ Q \circ D^2 f_{\varphi_{1}, \cdots, \varphi_m}(u)\right) \\
    =& \sum_{\ell = 1}^\infty \left< A(t) \circ Q \circ D^2 f_{\varphi_{1}, \cdots, \varphi_m}(u) (Q^{1/2}\varphi_\ell),  \, 
   Q^{1/2}\varphi_\ell 
    \right>_{\mathcal{H}_Q} \\ 
    =& \sum_{\ell = 1}^\infty \left\langle A(t) \circ Q \circ \left( 
    \sum_{i, j = 1}^m (\partial_{ij}^2f)(\varphi_1(u), \cdots, \varphi_m(u)) \left\langle \varphi_j, Q^{1/2} \varphi_\ell \right\rangle_{\mathcal{H}} \varphi_i 
    \right), \, Q^{1/2}\varphi_\ell \right\rangle_{\mathcal{H}_Q} \tag{\lemref{Lemma}}
    \\
    =& \sum_{i, j = 1}^m (\partial_{ij}^2 f)(\varphi_1(u), \cdots, \varphi_m(u)) \left[
    \sum_{\ell = 1}^\infty \left<
    A(t) \circ Q(\varphi_i), Q^{1/2} \varphi_\ell 
    \right>_{\mathcal{H}_Q} \left<
    \varphi_j, Q^{1/2} \varphi_\ell 
    \right>_{\mathcal{H}}
    \right]
    \\ 
    =& \sum_{i, j = 1}^m (\partial_{ij}^2 f)(\varphi_1(u), \cdots, \varphi_m(u)) \left[
    \sum_{\ell = 1}^\infty \left<
    A(t) \circ Q(\varphi_i), Q^{1/2} \varphi_\ell 
    \right>_{\mathcal{H}_Q} \left< Q^{1/2}
    \varphi_j, Q^{1/2} \circ Q^{1/2} \varphi_\ell 
    \right>_{\mathcal{H}_Q}
    \right]
    \\ 
    =& \sum_{i, j = 1}^m (\partial_{ij}^2 f)(\varphi_1(u), \cdots, \varphi_m(u)) \left[
    \sum_{\ell = 1}^\infty \left<
    A(t) \circ Q(\varphi_i), Q^{1/2} \varphi_\ell 
    \right>_{\mathcal{H}_Q} \left< Q
    \varphi_j,  Q^{1/2} \varphi_\ell 
    \right>_{\mathcal{H}_Q}
    \right] \tag{$Q^{1/2}$ is self-adjoint} \\
    =& \sum_{i, j = 1}^m (\partial_{ij}^2  f)(\varphi_1(u), \cdots, \varphi_m(u)) \left<
    A(t) \circ Q (\varphi_i), Q(\varphi_j)
    \right>_{\mathcal{H}_Q} 
    \\
    =& \sum_{i =1}^m \left[
    \sum_{j = 1}^m \partial_j \left( 
    \partial_i f (\varphi_1(u), \cdots, \varphi_m(u)) 
    \right) \left<A(t) \circ Q( \varphi_i), Q(\varphi_j)\right>_{\mathcal{H}_Q} 
    \right]\\
    =& \sum_{i = 1}^m \left<
    D(\partial_i f_{\varphi_{1,\cdots, m}}) (u), A(t) \circ Q(\varphi_i)
    \right>_{\mathcal{H}_Q}. \tag{Definition of the Fr\'echet derivative}
\end{align*}
\endgroup
Now, from the preceding chain of equalities, we write 
\begingroup
\allowdisplaybreaks
\begin{align*}
    &\int_{\mathcal{H}}  \operatorname{Tr}_{\mathcal{H}_Q} \left(
    A(t) \circ Q \circ D^2 f_{\varphi_{1}, \cdots, \varphi_m}(u)
    \right) \mu_t(du) \\
    =&\sum_{i = 1}^m \int_\mathcal{H}   \left<
    D(\partial_i f_{\varphi_{1,\cdots, m}}) (u), A(t) \circ Q(\varphi_i)
    \right>_{\mathcal{H}_Q} \mu_t(du) \\
    =& -\sum_{i = 1}^m \int_\mathcal{H} \partial_{i}  f_{\varphi_{1}, \cdots, \varphi_m}(u) \, 
    \rho_{A(t) \circ Q(\varphi_i)}^{\mu_t}(u) \, \mu_t(du) \tag{Integration-by-parts} \\
    =& -\sum_{i = 1}^m \int_\mathcal{H} \partial_{i}  f_{\varphi_{1}, \cdots, \varphi_m}(u) \left<
    \rho_{\mathcal{H}_Q}^{\mu_t}(u), A(t) \circ Q(\varphi_i)
    \right>_{\mathcal{H}_Q} \mu_t(du) \tag{Definition of the logarithmic gradient} \\
    =& -\sum_{i = 1}^m \int_\mathcal{H} \partial_i f_{\varphi_{1}, \cdots, \varphi_m}(u) \left< 
    A(t) \rho_{\mathcal{H}_Q}^{\mu_t} (u), Q(\varphi_i)
    \right>_{\mathcal{H}_Q }\mu_t(du), 
\end{align*}
\endgroup 
which eventually proves the claim \eqref{eq:claim}, and hence the theorem. \end{proof}

\begin{proof}[Proof of \lemref{Lemma}] From the linearity of inner product, note that 
\begin{align*}
    f_{\varphi_1, \cdots, \varphi_m} (u + h) &= f(\langle \varphi_1, u + h\rangle, \cdots, \langle \varphi_m, u + h \rangle) \\
    &= f(\langle\varphi_1, u \rangle + \varepsilon_1, \cdots, \langle \varphi_m,  u \rangle +\varepsilon_m), 
\end{align*}
where $\varepsilon_i = \langle \varphi_i, h \rangle. $ From the Cauchy-Schwwarz inequality, it is clear that $\varepsilon_i \to 0$ as $\|h\| \to 0$ for each $i = 1,2, \cdots, m$. Because $f \in \mathcal{C}_0^\infty(\RR^m)$ is smooth, it follows from the Taylor expansion of $f$ applied to the RHS of the preceding equality that 
\begin{align*}
    f_{\varphi_1, \cdots, \varphi_m}& (u + h) \\= &f(\langle \varphi_1, u \rangle, \cdots, \langle \varphi_m, u \rangle) + \sum_{i = 1}^m (\partial_i f)(\langle \varphi_1, u \rangle, \cdots, \langle \varphi_m, u \rangle) \langle \varphi_i, h \rangle+ o(\|h\|). 
\end{align*}
Therefore, it follows that 
\begin{align*}
    \partial_h f_{\varphi_1, \cdots, \varphi_m}(u) =
    \sum_{i = 1}^m (\partial_i f)(\langle \varphi_1, u \rangle, \cdots, \langle \varphi_m, u \rangle) \langle \varphi_i, h \rangle. 
\end{align*}
Because $\partial_h f_{\varphi_1, \cdots, \varphi_m}(u) = \langle Df_{\varphi_1, \cdots, \varphi_m}(u), h \rangle$, we conclude that 
\begin{align*}
    Df_{\varphi_1, \cdots, \varphi_m} (u) = \sum_{i = 1}^m (\partial_i f)(\langle \varphi_1, u \rangle, \cdots, \langle \varphi_m, u \rangle) \varphi_i. 
\end{align*}
Repeating the same method, one observes that 
\begin{align*}
      Df_{\varphi_1, \cdots, \varphi_m}&(u + h)\\
      =&   Df_{\varphi_1, \cdots, \varphi_m}(u) + \sum_{i, j = 1}^m (\partial_{ij}^2 f) (\langle \varphi_1, u \rangle, \cdots, \langle \varphi_m, u \rangle) \varphi_i \langle \varphi_j, h \rangle + o(\|h\|), 
\end{align*}
from which the second statement is deduced. This completes the proof.
\end{proof}

%% file: section/A03.tex
\newpage


\section{Experimental details}
\label{app:Experiments}

\subsection{Training and sampling}

\paragraph{Training.} We approximate $\rho_{\mathcal{H}_Q}^{\mu_t}(u)$ via a Fourier neural operator $S_\theta(t, u)$ parametrized by $\theta$. The training of $S_\theta$ is done via the score-matching objective~\citep{vincent2011connection, sohl2015deep}: 
\begin{align*}
    \minimize_{\theta} \quad \mathcal{L}(\theta) = \int_0^T \expectation_{X_0 \sim \PP_0} \left[ 
    \expectation_{X_t \sim \mu_{t|X_0}} \left[ \left\| 
    S_\theta(t, X_t) - \rho_{\mathcal{H}_Q}^{\mu_{t|X_0}}(X_t) 
    \right\|^2\right]
    \right], 
\end{align*}
where $\mu_{t|X_0}$ is the conditional measure of $X_t$ given $X_0$. We use the variance-preserving SDE~\citep{song2021score} in $\mathcal{H}$ of the form 
\begin{align*}
    dX_t^\rightarrow = -\frac{\alpha(t)}{2} X_t^\rightarrow + \sqrt{\alpha(t)} X_t^\rightarrow \, dW_t, \quad X_0 \sim \PP_0= \PP_{\mathrm{data}} 
\end{align*}
where $(W_t)_{t \ge 0}$ is a $Q$-Wiener process in $\mathcal{H}$.

\paragraph{Sampling.} Once $S_\theta$ is learned, we sample synthetic data by plugging in $S_\theta(T -t, Y_{T - t}^{\leftarrow})$ in place of 
$\rho_{\mathcal{H}_Q}^{\mu_{T _ t}}(Y_{T - t}^{\leftarrow})$ in our probability-flow ODE (Theorem~\ref{thm:prob-flow-ode}) as
\begin{align}
    dY_{T - t}^{\leftarrow} = \frac{\alpha(T - t)}{2}\left[ Y_{T - t}^{\leftarrow} + \rho_{\mathcal{H}_Q}^{\mu_{T - t}}(Y^{\leftarrow}_{T - t})
    \right] \, dt, \quad Y_0\sim \mathcal{N}(0, Q),
    \label{eq:NonPLP}
\end{align}
and run the following plug-and-play ODE 
\begin{align}
    dY_{T - t}^{\leftarrow} = \frac{\alpha(T -t)}{2}\left[ Y_{T - t}^{\leftarrow} + S_\theta(T - t, Y^{\leftarrow}_{T - t})
    \right] \, dt, \quad Y_0\sim \mathcal{N}(0, Q).
    \label{eq:PLP}
\end{align}
In every example in this work, we utilize Euler's method to solve the preceding plug-and-play ODE (\eqref{eq:PLP}).

\vspace{5mm} 

\subsection{Approximation of $\mathcal{N}(0, Q)$} 

\paragraph{1D~(Quadratic) function generation.}

For one-dimensional function generation task, we let $k(\cdot, \cdot) : [a, b] \times [a, b]  \to \RR$ be a positive-definite kernel. We let $Q$ be the integral operator on $L^2([a, b])$ corresponding to $k$. 
Let $\mathcal{D} = \{x_1, \cdots, x_N\}$ be the fixed grid where functions are evaluated. Define the Gram matrix $K \in \operatorname{Mat}_N(\RR)$ by $K_{ij} = k(x_i, x_j)$, and let $K = \Phi D \Phi^\intercal$ be the eigen-decomposition of $K$. As in \cite{ baker1979numerical, phillips2022spectral, lim2023score}, we generate a random noise $W \sim \mathcal{N}(0, Q)$ by 
\begin{align*}
    W = Z D^{1/2} \Phi^\intercal, \quad Z \sim \mathcal{N}(0, id_N). 
\end{align*}
We choose our $k$ to be the Gaussian RBF kernel of the form 
\begin{align*}
    k(x_1, x_2) = \texttt{gain} \, e^{-|x_1 - x_2|^2/\texttt{len}^2}, \quad x_1, x_2 \in [a, b].
\end{align*}
The hyperparameters \texttt{gain} and \texttt{len} are called the gain parameter and the length parameter, respectively.

\paragraph{PDE problems.}

For PDE solution generation tasks~(reaction-diffusion equation and time-evolving heat equation), we use the Bessel prior $\mathcal{N}(0, (\gamma - \Delta)^{-s})$, which is introduced in \cite{hagemann2023multilevel}, as our noise $\mathcal{N}(0, Q)$ in the ambient Hilbert space $\mathcal{H}$. Sampling from $\mathcal{N}(0, (\gamma - \Delta)^{-s})$ is done by computing
\begin{align*}
    W = \mathrm{FFT}^{-1}\left(
    \lambda(\gamma - \Delta)^{-s/2}) \, \odot \, \mathrm{FFT}(Z)
    \right), \quad Z \sim \mathcal{N}(0, id_{N^2}), 
\end{align*}
where $\lambda((\gamma - \Delta)^{-s/2})$ is the vector whose entries consist of eigenvalues of $(\gamma - \Delta)^{-s/2}$, $N$ is the resolution of samples, $\odot$ denotes the entry-wise product, and $\mathrm{FFT}$ and $\mathrm{FFT}^{-1}$ denotes the Fast Fourier Transform and its inverse transform, respectively. 
The hyperparameter $\gamma > 0$ is called the \texttt{scale} parameter, and $s > 0$ is called the \texttt{power} parameter.

\vspace{5mm}

\subsection{Implementational details} 

In this subsection, we provide details regarding the architecture and training details of the infinite-dimensional diffusion models used in our experiments.

\paragraph{1D~(Quadratic) function generation.}

For the one-dimensional~(\texttt{Quadratic}) function generation task, we use a modified version of Fourier Neural Operator~\citep{li2020fourier} that is proposed in \cite{lim2023score}. Here, we use the pre-trained checkpoint by \cite{lim2023score}{ without additional training}\footnote{Official code repository: \url{https://github.com/KU-LIM-Lab/hdm-official/}}. Table~\ref{table:quadratic-exp} lists up detailed architectural design used in the 1D function generation experiment.

\begin{table}[H]
\small
\caption{Architectural details for 1D~(\texttt{Quadratic}) function generation}
\label{table:quadratic-exp}
\centering
\begin{tabular}{lll}
\toprule
\textbf{Architecture} & Base channels            & 256  \\
        & \# of ResBlocks per stage & 4  \\
        & Lifting channels & 256 \\ 
        & Projection channels & 256 \\ 
        & \# of modes  &[100] \\
        & Activation function       & Gelu \\
        & Normalization & GroupNorm \\ 
\midrule
\textbf{Diffusion}   & Noise schedule            & Cosine  \\
        & \# timesteps & 1000 \\ 
                           & $\log( \alpha_0^2 / \sigma_0^2 )$ & 10  \\
                           & $\log( \alpha_1^2 / \sigma_1^2 )$ & -10  \\
        & Length parameter & 0.8 \\ 
        & Gain parameter & 1.0 \\
\bottomrule
\end{tabular}
\end{table}

\paragraph{2D Reaction-diffusion equation.} For the two-dimensional reaction-diffusion equation problem, we use a two-dimensional Fourier Neural Operator. We follow the architectural detail used in \cite{hagemann2023multilevel}\footnote{Official code repository: \url{https://github.com/PaulLyonel/multilevelDiff/}}, which is based on the official implementation of Neural Operators~\citep{kovachki2021neural, kossaifi2024neural}\footnote{Official code repository: \url{https://github.com/neuraloperator/neuraloperator/}}. Table~\ref{table:reacdiff-exp} lists up detailed architectural design used in the reaction-diffusion equation experiment.

\begin{table}[H]
\small
\caption{Implementational details for 2D~(reaction-diffusion) experiment}
\label{table:reacdiff-exp}
\centering
\begin{tabular}{lll}
\toprule
\textbf{Architecture} & Base channels            & 32  \\
        & \# of ResBlocks per stage & 4  \\
        & Lifting channels & 32 \\ 
        & Projection channels & 128 \\ 
        & \# of modes  & [12, 12] \\
        & Activation function       & Gelu \\
        & Normalization & None \\ 
\midrule
\textbf{Diffusion}   & Noise schedule            & Cosine  \\
        & \# timesteps & 1000 \\ 
                           & $\log( \alpha_0^2 / \sigma_0^2 )$ & 10  \\
                           & $\log( \alpha_1^2 / \sigma_1^2 )$ & -10  \\
        & Scale parameter & 8 \\
        & Power parameter & 0.55 \\
\midrule
\textbf{Training}   
    & Optimizer     & Adam, $\beta_1 =0.9$, $\beta_2 = 0.999$ \\
    & Learning rate & 0.001 \\ 
    & \# epochs & 200 \\ 
    & Batch size & 8 \\
\bottomrule
\end{tabular}
\end{table}

\paragraph{2D Heat equation.} For the two-dimensional heat equation, we discretize time-evolving solutions of the heat equation on a three-dimensional domain: two dimensions for spacial and one for time. We use a three-dimensional Fourier Neural Operator~\citep{li2020fourier, kovachki2021neural, kossaifi2024neural}.
Table~\ref{table:heat-exp} lists up implementational details used in the time-evolving heat equation experiment.

\begin{table}[H]
\small
\caption{Implementational details for time-evolving heat equation experiment}
\label{table:heat-exp}
\centering
\begin{tabular}{lll}
\toprule
\textbf{Architecture} & Base channels            & 32  \\
        & \# of ResBlocks per stage & 4  \\
        & Lifting channels & 32 \\ 
        & Projection channels & 128 \\ 
        & \# of modes  & [12, 12, 12] \\
        & Activation function       & Gelu \\
        & Normalization & None \\ 
\midrule
\textbf{Diffusion}   & Noise schedule            & Cosine  \\
        & \# timesteps & 1000 \\ 
                           & $\log( \alpha_0^2 / \sigma_0^2 )$ & 10  \\
                           & $\log( \alpha_1^2 / \sigma_1^2 )$ & -10  \\
        & Scale parameter & 8 \\ 
        & Power parameter & 0.55\\ 
\midrule
\textbf{Training}   
    & Optimizer     & Adam, $\beta_1 =0.9$, $\beta_2 = 0.999$ \\
    & Learning rate & 0.001 \\ 
    & \# epochs & 200 \\ 
    & Batch size & 4 \\
\bottomrule
\end{tabular}
\end{table}